\documentclass{article}

\usepackage{microtype}
\usepackage{graphicx}
\usepackage{subfigure}
\usepackage{booktabs} 

\usepackage{hyperref}

\usepackage[accepted]{icml2025}

\usepackage{amsmath}
\usepackage{amssymb}
\usepackage{mathtools}
\usepackage{amsthm}
\usepackage{mathrsfs}
\usepackage[capitalize,noabbrev]{cleveref}

\theoremstyle{plain}
\newtheorem{theorem}{Theorem}[section]
\newtheorem{proposition}[theorem]{Proposition}

\newtheorem{corollary}[theorem]{Corollary}
\theoremstyle{definition}

\theoremstyle{remark}
\newtheorem{remark}[theorem]{Remark}

\usepackage[textsize=tiny]{todonotes}

\icmltitlerunning{Finite-Horizon Input–Output Dynamics of Minibatch Perturbations in AdamW}

\begin{document}

\twocolumn[
\icmltitle{Finite-Horizon Input–Output Dynamics of Minibatch Perturbations in AdamW}



\icmlsetsymbol{equal}{*}

\begin{icmlauthorlist}
\icmlauthor{Kang Liu}{xjtu}
\icmlauthor{Suyan Li}{nus}
\end{icmlauthorlist}

\icmlaffiliation{xjtu}{School of Future Technology, Xi'an Jiaotong University, Xi'an, China, 710049}
\icmlaffiliation{nus}{Department of Electrical and Computer Engineering, National University of Singapore, Singapore, 119077}
\icmlcorrespondingauthor{Kang Liu}{kanyo@foxmail.com}

\icmlkeywords{Optimizer dynamics, Minibatch perturbation, Finite-Horizon analysis, AdamW}

\vskip 0.3in
]



\printAffiliationsAndNotice{}  

\begin{abstract}
A minibatch can influence training beyond the update at which it is observed because AdamW stores past gradient information in its optimizer states. We study this delayed effect through paired trajectories that differ only in one gradient update and share the same subsequent training sequence. We formulate AdamW as a finite-horizon input--state--output (ISO) system whose state contains the model parameters and first- and second-moment estimates. Linearizing the joint dynamics yields a signed response operator that maps a localized gradient perturbation to its future loss effects, revealing how optimizer memory shapes their magnitude, timing, and sign.
We further derive an exact multistep error decomposition and establish first-order finite-horizon accuracy under local smoothness and controlled activation switching. Experiments validate the response mechanism and optimizer-state effects, while repeated-future analyses reveal substantial prospective structure in delayed influence that can be partially recovered from ISO approximations. Code is available at \url{https://github.com/Kanyooo/Loss_ISO}.
\end{abstract}

\section{Introduction}
\label{sec:introduction}

Modern neural networks are trained through a sequence of stochastic minibatch
updates. With adaptive optimizers such as AdamW, however, the effect of a
minibatch is not confined to the step at which it is observed. Its gradient
changes not only the model parameters, but also the first- and second-moment
estimates that determine subsequent updates. As a result, the effect of one
minibatch can persist for several training steps and may become visible in the
loss only later. This temporal dependence is particularly relevant to transient
training instabilities, including loss spikes. Sharp local geometry and
large-step training dynamics have been linked to transient loss excursions
\citep{zhu2024catapults}. Small-scale proxies further show that local training
statistics can anticipate large-scale Transformer instabilities
\citep{wortsman2024proxies}. More recent work attributes another class of spikes
to abnormal stochastic gradients and motivates spike-aware momentum treatment
\citep{huang2025spam}.

Existing analyses characterize Adam through local geometry, adaptive
preconditioning, and optimizer dynamics
\citep{das2024preconditioning,ahn2024adam}. Recent influence studies further
show that training-data effects depend on their position along the optimization
trajectory \citep{wang2025temporal}, and that AdamW-aware attribution must
propagate perturbations through both parameter and moment states
\citep{deng2026faithful}. Optimizer memory can also make minibatch ordering a
first-order source of finite-window variation
\citep{sweeney2026optimizerMemory}. \textit{These results establish that optimizer
state and trajectory matter, but they do not provide a signed,
horizon-resolved account of how a localized minibatch perturbation is stored,
propagated, and expressed in future loss.} 

We study the delayed effect of minibatch perturbations in AdamW by comparing
paired training trajectories that differ only at one update and share the same
subsequent training sequence. This construction isolates how a localized
perturbation evolves over a finite future horizon.

\begin{figure*}[t]
    \centering
    \includegraphics[width=0.9\textwidth]{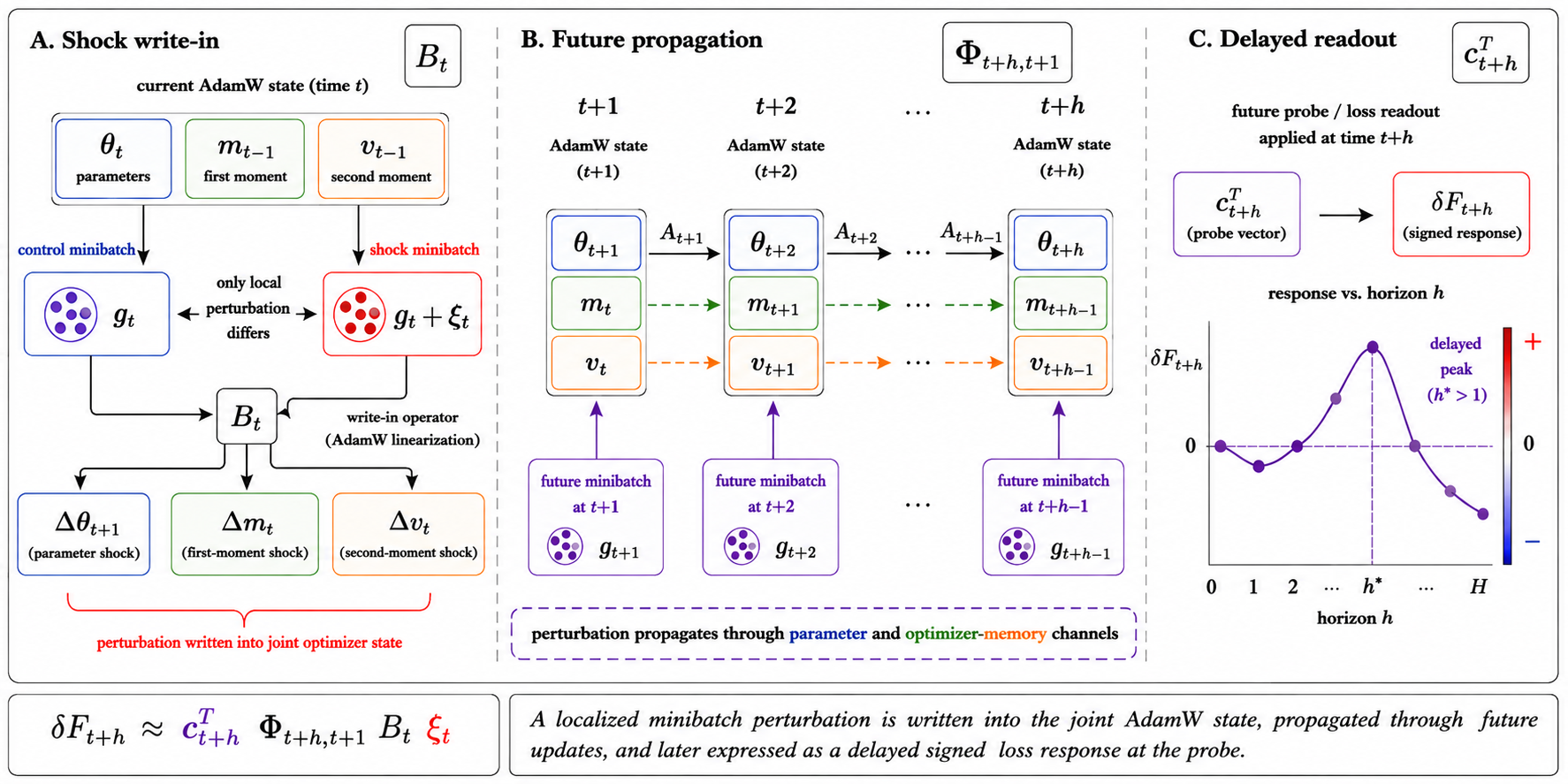}
    \caption{Finite-horizon ISO view of delayed minibatch influence in AdamW.}
    \label{fig:adamw_mechanism}
\end{figure*}

The central idea is to formulate AdamW as a finite-horizon
input--state--output (ISO) system, with the model parameters and optimizer
moments forming the joint state. We derive a tangent model that maps an
initial gradient perturbation to its future loss response through the
intervening AdamW dynamics, revealing how optimizer memory shapes the
magnitude, timing, and sign of delayed effects. We further characterize the
finite-horizon approximation error, including nonlinear dynamics and
activation-pattern changes. Beyond pathwise analysis, we ask whether delayed
influence retains structure before the future training sequence is realized.
Repeated-future experiments show substantial prospective structure, while its
recoverability from present-time ISO approximations depends on the underlying
dynamics. Figure~\ref{fig:adamw_mechanism} summarizes the ISO mechanism.

Our contributions are threefold:
\begin{enumerate}
    \item
    We formulate localized minibatch influence as a signed, finite-horizon response under paired AdamW trajectories, isolating the effect of a single gradient perturbation under a shared future realization.

    \item
    We derive a joint parameter--moment ISO operator that decomposes the response
    into perturbation write-in, optimizer-state propagation, and loss readout, thereby characterizing how AdamW memory can delay or transiently amplify a minibatch's effect.

    \item
    We establish an exact multistep error decomposition that separates
    write-in nonlinearity, smooth propagation error, and activation switching,
    and prove fixed-horizon first-order accuracy.
\end{enumerate}

\section{Related Work}
\label{sec:related_work}

Training-data influence has gradually moved from static, endpoint-based
attribution toward trajectory-aware descriptions of how individual training
examples affect learning. Classical influence functions characterize
infinitesimal reweighting around a trained solution
\citep{koh2017understanding}, while subsequent work has improved their
scalability and applicability to modern deep networks and large models
\citep{schioppa2022scaling,grosse2023studying,park2023trak,
kwon2024datainf,xia2024less}. A parallel line instead follows the optimization
path, from retracing SGD updates and propagating hypergradients
\citep{hara2019data,pruthi2020tracin,chen2021hydra}
to approximate unrolling and trajectory-specific influence
\citep{bae2024source,wang2025temporal}. Recent work further shows that faithful
trajectory attribution under AdamW requires accounting for its parameter and
moment states \citep{deng2026faithful}. These developments establish that
training order and intermediate optimization states matter. The object studied
here is more local and dynamical: we perturb one realized minibatch gradient and track its signed response over a finite horizon, making the temporal response itself the quantity of interest.

For AdamW, this distinction is important because the optimizer carries
information across updates through its first- and second-moment states.
Existing work has characterized Adam and AdamW through convergence,
preconditioning, implicit geometry, and dynamical behavior
\citep{kingma2015adam,reddi2018convergence,loshchilov2019decoupled,
ma2022qualitative,ahn2024adam,lin2024square,xie2024adamw}.
Closely related studies of training instability show that adaptive
preconditioning, gradient statistics, and optimizer-state dynamics can shape
short-time behavior and transient loss spikes
\citep{wortsman2024proxies,zhu2024catapults,huang2025spam,
wang2025adagc,bai2025adaptive}. Optimizer memory can also make minibatch
ordering a first-order source of finite-window variation
\citep{sweeney2026optimizerMemory}. These results motivate treating the
optimizer state as part of the perturbation dynamics rather than as an
implementation detail. Our formulation treats parameters and AdamW moments as a joint state, explicitly separating perturbation write-in, state propagation, and loss readout. This finite-horizon ISO view provides a signed, horizon-resolved account of how a localized minibatch perturbation is stored in optimizer state and later expressed in future loss.

\section{Finite-Horizon Minibatch Influence}
\label{sec:finite_horizon}

\subsection{Overview and Time Convention}
\label{sec:fh_overview}

We model the effect of a gradient perturbation as a finite-horizon
input--state--output process:
\[
    \xi_t
    \xrightarrow{\;B_t\;}
    \delta x_{t+1}
    \xrightarrow{\;\Phi_{t+h,t+1}\;}
    \delta x_{t+h}
    \xrightarrow{\;c_{t+h}^{\top}\;}
    \delta F_{t+h}.
\]
Here, $\xi_t$ denotes the perturbation applied at step $t$, $B_t$ maps it into
the joint AdamW state, $\Phi_{t+h,t+1}$ propagates the resulting state
deviation through subsequent updates, and $c_{t+h}^{\top}$ maps the propagated
deviation to the probe-loss response.

Throughout, $x_s$ denotes the joint state immediately before minibatch $s$ is
processed. Since the perturbation is applied during update $t$, horizon $h=1$
corresponds to the first post-perturbation state $x_{t+1}$.

\subsection{Paired Trajectories and Joint AdamW State}
\label{sec:paired_system}

For minibatch $\mathcal B_s$, define
\(
    g_s(\theta)
    :=
    \nabla_{\theta}\ell(\theta;\mathcal B_s)
    \in\mathbb R^d.
\)
Immediately before processing minibatch $s$, the joint AdamW state is
\(
    x_s
    :=
    [
        \theta_s,
        m_{s-1},
        v_{s-1}
    ]^\top
    \in\mathbb R^{3d},
\)
where $\theta_s$ is the model parameter, $m_{s-1}$ is the first-moment state,
and $v_{s-1}$ is the second-moment state. Under AdamW \citep{loshchilov2019decoupled}, the first- and second-moment
estimates evolve as
\[
\begin{aligned}
    m_s
    &=
    \beta_1m_{s-1}
    +(1-\beta_1)g_s(\theta_s),\\
    v_s
    &=
    \beta_2v_{s-1}
    +(1-\beta_2)g_s(\theta_s)^{\odot 2}.
\end{aligned}
\]
Let
$\rho_{1,s}:=1-\beta_1^s$ and $\rho_{2,s}:=1-\beta_2^s$,
and define the corrected moments
$\widehat m_s:=m_s/\rho_{1,s}$ and $\widehat v_s:=v_s/\rho_{2,s}$.
The parameter update is written as
\[
    \theta_{s+1}
    =
    D_s\theta_s+q_s(m_s,v_s),
    ~~
    D_s:=I-\eta_s\Lambda_s,
\]
where $\Lambda_s$ denotes decoupled weight decay and
\[
    q_s(m,v)
    :=
    -\eta_s
    \left(
        m/{\rho_{1,s}}
    \right)
    \oslash
    \left(
        \sqrt{{v}/{\rho_{2,s}}}
        +\epsilon\mathbf 1
    \right)
\]
is evaluated coordinatewise.

\paragraph{Paired trajectories.}
We compare a control trajectory and a shock trajectory initialized from the
same state:
$x_t^{\mathrm s}=x_t^{\mathrm c}=x_t$.
At the shock step, their gradients satisfy
$g_t^{\mathrm s}=g_t^{\mathrm c}+\xi_t$,
where $\xi_t$ is the realized minibatch-gradient perturbation. From step
$t+1$ onward, the two trajectories use the same minibatches, stochastic
realizations, learning-rate schedule, and optimizer configuration.

Define the paired state deviation
$\delta x_s:=x_s^{\mathrm s}-x_s^{\mathrm c}$.
Let $\{F_s:\mathbb R^d\rightarrow\mathbb R\}$ be a sequence of probe
objectives shared by the paired trajectories, and define
\[
    \delta F_s
    :=
    F_s(\theta_s^{\mathrm s})
    -
    F_s(\theta_s^{\mathrm c}).
\]
A fixed probe corresponds to the special case $F_s\equiv F$. Alternatively,
taking
\[
    F_s(\theta)
    :=
    \ell(\theta;\mathcal B_s)
\]
gives the response of the realized future minibatch loss under the shared
future sequence. Because the probe and future random sequence are shared,
$\delta F_s$ isolates the pathwise effect of the gradient difference
introduced at step $t$. The response is signed:
$\delta F_s>0$ means that the shock raises the probe loss relative to the
control, whereas $\delta F_s<0$ means that it lowers the probe loss.

The control gradient $g_t^{\mathrm c}$ may be the gradient of another
realized minibatch, as in the paired experimental protocol. Alternatively,
taking it to be the conditional mean gradient,
\[
    \bar g_t
    :=
    \mathbb E[g_t\mid\mathcal F_t],
    ~~
    \xi_t:=g_t-\bar g_t,
\]
gives a conditional stochastic interpretation of the perturbation, where
$\mathcal F_t$ denotes the training history available before sampling the
minibatch at step $t$.

\subsection{Minibatch Shock Write-In}
\label{sec:shock_writein}

We first derive how the perturbation $\xi_t$ enters the three blocks of the
joint AdamW state. Unless stated otherwise, all quantities in this subsection
are evaluated along the control trajectory. Define the sensitivities of the
adaptive parameter update to the moment states:
\[
    M_s
    :=
    \frac{\partial q_s}{\partial m_s},
    ~~
    V_s
    :=
    \frac{\partial q_s}{\partial v_s}.
\]
Writing
$d_s:=\sqrt{\widehat v_s}+\epsilon\mathbf 1$,
these sensitivities are diagonal:
{\small
\begin{align*}
    M_s
    &=
    -\frac{\eta_s}{\rho_{1,s}}
    \operatorname{Diag}(d_s^{-1}),\\
    V_s
    &=
    \eta_s
    \operatorname{Diag}
    \left(
        \frac{
            \widehat m_s
        }{
            2\rho_{2,s}
            \sqrt{\widehat v_s}
            \odot d_s^{\odot 2}
        }
    \right).
\end{align*}
}
Derivatives with respect to the second-moment state are evaluated at
coordinates where $\widehat v_{s,i}>0$. The matrix $M_s$ is negative
diagonal, while the sign of each diagonal entry of $V_s$ follows the
corresponding first-moment coordinate. Define the effective
gradient-to-parameter Jacobian
\[
    \Gamma_s
    :=
    (1-\beta_1)M_s
    +
    2(1-\beta_2)
    V_s\operatorname{Diag}(g_s^{\mathrm c}).
\]
Its two terms correspond to the paths
$g\rightarrow m\rightarrow\theta$ and
$g\rightarrow v\rightarrow\theta$, respectively.

At the shock step, direct subtraction of the two moment updates gives
\begin{align*}
    \delta m_t
    &=
    (1-\beta_1)\xi_t,\\
    \delta v_t
    &=
    2(1-\beta_2)
    \operatorname{Diag}(g_t^{\mathrm c})\xi_t
    +(1-\beta_2)\xi_t^{\odot 2}.
\end{align*}
Thus, the first-moment write-in is exactly linear, whereas the second-moment
write-in contains both linear and quadratic components.

Define the composite gradient-to-parameter map
\[
    \psi_t(g)
    :=
    q_t\Bigl(
        \beta_1m_{t-1}+(1-\beta_1)g,\;
        \beta_2v_{t-1}+(1-\beta_2)g^{\odot2}
    \Bigr).
\]
Then
$\Gamma_t=D\psi_t(g_t^{\mathrm c})$,
and
\[
    \delta\theta_{t+1}
    =
    \Gamma_t\xi_t+r_{\theta,t}(\xi_t),
\]
where
\[
    r_{\theta,t}(\xi_t)
    :=
    \psi_t(g_t^{\mathrm c}+\xi_t)
    -
    \psi_t(g_t^{\mathrm c})
    -
    \Gamma_t\xi_t.
\]
If $D\psi_t$ is $\kappa_{\theta,t}$-Lipschitz along the segment joining
$g_t^{\mathrm c}$ and $g_t^{\mathrm c}+\xi_t$, then
\[
    \|r_{\theta,t}(\xi_t)\|
    \leq
    \frac{\kappa_{\theta,t}}{2}
    \|\xi_t\|^2.
\]

Stacking the three state blocks yields
\[
    \delta x_{t+1}
    =
    B_t\xi_t+r_{B,t}(\xi_t),
\]
where
{\footnotesize
\begin{equation}
    B_t
    :=
    \begin{bmatrix}
        \Gamma_t\\[1mm]
        (1-\beta_1)I\\[1mm]
        2(1-\beta_2)
        \operatorname{Diag}(g_t^{\mathrm c})
    \end{bmatrix},
    ~~
    r_{B,t}(\xi_t)
    :=
    \begin{bmatrix}
        r_{\theta,t}(\xi_t)\\[1mm]
        0\\[1mm]
        (1-\beta_2)\xi_t^{\odot 2}
    \end{bmatrix}.
    \label{eq:writein_remainder}
\end{equation}
}

The remainder contains two distinct nonlinear effects: the intrinsic
quadratic write-in to the second-moment state and the nonlinearity of the
adaptive parameter update. The current weight-decay term does not appear in
$B_t$ because the two trajectories share the same pre-shock parameter
$\theta_t$; weight decay instead enters subsequent transitions through
$D_s=I-\eta_s\Lambda_s$.

\subsection{Joint-State Propagation and Loss Readout}
\label{sec:joint_state_propagation}

Let $f_s$ denote the exact state transition under future minibatch
$\mathcal B_s$. For a smooth network, set $\widetilde f_s:=f_s$. For a
piecewise-smooth network, let $\widetilde f_s$ denote the smooth extension
associated with the activation pattern of the control trajectory at
$x_s^{\mathrm c}$. The tangent transition matrix is
\[
    A_s:=D\widetilde f_s(x_s^{\mathrm c}).
\]

Although the paired trajectories use the same future minibatch, their gradients
generally differ once their parameters diverge. Let $\widetilde g_s$ denote the
gradient map within the control-region smooth extension, and define
\[
    H_s^{\mathrm{tr}}
    :=
    D\widetilde g_s(\theta_s^{\mathrm c}).
\]
When the loss is twice differentiable in this region,
\[
    H_s^{\mathrm{tr}}
    =
    \nabla_{\theta}^2
    \ell(\theta_s^{\mathrm c};\mathcal B_s).
\]
The corresponding first-order gradient variation is
\[
    \widetilde g_s
    (\theta_s^{\mathrm c}+\delta\theta_s)
    -
    \widetilde g_s(\theta_s^{\mathrm c})
    =
    H_s^{\mathrm{tr}}\delta\theta_s
    +
    o(\|\delta\theta_s\|).
\]

Define
\[
    S_s
    :=
    2(1-\beta_2)
    \operatorname{Diag}(g_s^{\mathrm c})
    H_s^{\mathrm{tr}}.
\]
The tangent transition of the joint parameter--moment state is
\begin{equation}
    A_s
    =
    \begin{bmatrix}
        D_s+\Gamma_sH_s^{\mathrm{tr}}
        &
        \beta_1M_s
        &
        \beta_2V_s
        \\[1mm]
        (1-\beta_1)H_s^{\mathrm{tr}}
        &
        \beta_1I
        &
        0
        \\[1mm]
        S_s
        &
        0
        &
        \beta_2I
    \end{bmatrix}.
    \label{eq:joint_state_jacobian}
\end{equation}
The off-diagonal blocks capture the two feedback paths
\[
    \theta\rightarrow g\rightarrow m\rightarrow\theta,
    ~~
    \theta\rightarrow g^{\odot2}\rightarrow v\rightarrow\theta.
\]
Hence, the persistence parameters $\beta_1$ and $\beta_2$ describe memory
within the individual moment states but do not by themselves determine the
stability or finite-horizon gain of the coupled system.

For indices $a\geq b$, define the ordered propagator
\[
    \Phi_{a,b}
    :=
    \begin{cases}
        I,
        & a=b,\\[1mm]
        A_{a-1}A_{a-2}\cdots A_b,
        & a>b.
    \end{cases}
\]
In particular,
\[
    \Phi_{t+h,t+1}
    =
    A_{t+h-1}\cdots A_{t+1},
    ~~ h\geq2,
\]
with $\Phi_{t+1,t+1}=I$. The matrices $A_s$ vary along the future training
trajectory and generally do not commute.

Define the probe readout vector
\[
    c_s
    :=
    \begin{bmatrix}
        \nabla F_s(\theta_s^{\mathrm c})\\
        0\\
        0
    \end{bmatrix}
    \in\mathbb R^{3d}.
\]
The probe loss depends directly only on the parameter block, so a deviation
stored in the moment states affects the probe loss only after it is converted
into parameter motion.

\begin{theorem}[Finite-horizon directional response]
\label{thm:finite_horizon_response}
Consider the scaled perturbation
\[
    g_t^{\mathrm s}(\alpha)
    =
    g_t^{\mathrm c}+\alpha\xi_t.
\]
Suppose that the control trajectory lies in the interior of the smooth regions
used to define $\widetilde f_s$, that the corresponding AdamW transitions are
differentiable along the trajectory, and that $F_s$ is differentiable at the
relevant control states. In particular, the second-moment coordinates involved
in the derivatives satisfy $\widehat v_{s,i}^{\mathrm c}>0$.
Let $\delta x_{t+h}(\alpha)$ and $\delta F_{t+h}(\alpha)$ denote the resulting
paired state and probe-loss responses. Then, for every fixed $h\geq1$,
\begin{align*}
    \left.
    \frac{\mathrm d}{\mathrm d\alpha}
    \delta x_{t+h}(\alpha)
    \right|_{\alpha=0}
    &=
    \Phi_{t+h,t+1}B_t\xi_t,\\
    \left.
    \frac{\mathrm d}{\mathrm d\alpha}
    \delta F_{t+h}(\alpha)
    \right|_{\alpha=0}
    &=
    c_{t+h}^{\top}
    \Phi_{t+h,t+1}
    B_t\xi_t.
\end{align*}
Define the horizon-$h$ input--output operator
\[
    \mathcal G_{t,h}
    :=
    c_{t+h}^{\top}
    \Phi_{t+h,t+1}
    B_t.
\]
Then
\[
    \delta F_{t+h}(\alpha)
    =
    \alpha\mathcal G_{t,h}\xi_t
    +
    o(\alpha),
    ~~
    \alpha\rightarrow0.
\]
\end{theorem}

The complete proof is given in Appendix~\ref{app:fh_response}.
Theorem~\ref{thm:finite_horizon_response} gives the central ISO decomposition:
a perturbation first enters the joint AdamW state through $B_t$, is propagated
by the intervening dynamics through $\Phi_{t+h,t+1}$, and is finally observed
through the loss readout $c_{t+h}^{\top}$. The future response therefore
depends on the interaction of perturbation direction, optimizer dynamics, and
future loss sensitivity rather than on any one of these quantities alone.

\begin{remark}
A second-order analysis of the one-step response and the coordinate-invariance
property of the ISO operator are provided in
Appendix~\ref{app:one_step_output} and
Appendix~\ref{app:coordinate_invariance}, respectively.
\end{remark}

\paragraph{Finite-horizon response summaries.}
For a horizon $H$, we summarize the response by its largest magnitude,
its timing, and its sign:
\[
    M_{t,H}
    :=
    \max_{1\leq h\leq H}
    |\delta F_{t+h}|,
    ~~
    h_{t,H}^{\star}
    :=
    \min\arg\max_{1\leq h\leq H}
    |\delta F_{t+h}|,
\]
and
\[
    s_{t,H}^{\star}
    :=
    \operatorname{sign}
    \left(
        \delta F_{t+h_{t,H}^{\star}}
    \right).
\]
Their tangent counterparts are obtained by replacing
$\delta F_{t+h}$ with $\mathcal G_{t,h}\xi_t$. A delayed extremal response
corresponds to $h_{t,H}^{\star}>1$.

When only adverse loss increases are of interest, we additionally use
\[
    P_{t,H}^{+}
    :=
    \max_{1\leq h\leq H}
    [\delta F_{t+h}]_+,
    ~~
    [a]_+:=\max(a,0).
\]
This distinction is important for loss spikes: a large absolute response need
not correspond to an increase in loss.

\paragraph{Relation to loss spikes.}
When $F_{t+h}$ is chosen as the realized future minibatch loss, let
$T_{t+h}$ denote the corresponding spike threshold and suppose
\[
    F_{t+h}(\theta_{t+h}^{\mathrm c})
    \leq
    T_{t+h}.
\]
The perturbed trajectory crosses the threshold at horizon $h$ exactly when
\[
    \delta F_{t+h}
    >
    T_{t+h}
    -
    F_{t+h}(\theta_{t+h}^{\mathrm c}).
\]
Thus, a positive finite-horizon response contributes to a loss spike only when
it exceeds the remaining margin to the threshold.

\subsection{Delayed Expression Through AdamW Memory}
\label{sec:frozen_adamw_response}

To isolate the role of optimizer memory, consider a scalar frozen-coefficient
approximation of the joint AdamW dynamics over a short horizon:
\[
    z_{h+1}
    =
    \overline A z_h,
    ~~
    \overline A
    :=
    \begin{bmatrix}
        a & b_m & b_v\\
        d_m & \beta_1 & 0\\
        d_v & 0 & \beta_2
    \end{bmatrix},
\]
where
\[
    z_h
    :=
    \begin{bmatrix}
        \delta\theta_h,
        \delta m_{h-1},
        \delta v_{h-1}
    \end{bmatrix}^{\top}.
\]
The coefficients are the scalar counterparts of the blocks in
Equation~\eqref{eq:joint_state_jacobian}:
\begin{align*}
    a
    :=
    D+\Gamma H^{\mathrm{tr}},~~
    b_m
    :=
    \beta_1M,~~
    b_v
    :=
    \beta_2V,\\
    d_m
    :=
    (1-\beta_1)H^{\mathrm{tr}},~~
    d_v
    :=
    2(1-\beta_2)gH^{\mathrm{tr}}.
\end{align*}
Here, $b_m$ and $b_v$ convert moment-state deviations into parameter motion,
whereas $d_m$ and $d_v$ feed parameter-induced gradient changes back into the
two moment states.

To study how a perturbation stored in either memory channel becomes expressed
in the parameter state, let
\[
    e_\theta:=[1,0,0]^\top,~~
    e_m:=[0,1,0]^\top,~~
    e_v:=[0,0,1]^\top,
\]
and define
\[
    r_k(h)
    :=
    e_\theta^\top
    \overline A^{\,h-1}
    e_k,
    ~~
    k\in\{m,v\}.
\]
A nonzero scalar probe sensitivity can be applied afterward as an output
scaling; the analysis below concerns the magnitude and timing of the
memory-to-parameter response.

\begin{proposition}[Finite-horizon memory-channel response]
\label{prop:adamw_memory_response}
For $k\in\{m,v\}$,
\begin{align*}
    r_m(1)&=0,
    &
    r_m(2)&=b_m,
    &
    r_m(3)&=b_m(a+\beta_1),\\
    r_v(1)&=0,
    &
    r_v(2)&=b_v,
    &
    r_v(3)&=b_v(a+\beta_2),
\end{align*}
and
\begin{equation}
    r_k(4)
    =
    b_k
    \Bigl(
        a^2+a\beta_k+\beta_k^2
        +b_md_m+b_vd_v
    \Bigr).
    \label{eq:four_step_memory_response}
\end{equation}

If $b_k\neq0$ and
\begin{equation}
    |a+\beta_k|>1,
    \label{eq:two_step_amplification_condition}
\end{equation}
then
\[
    |r_k(3)|>|r_k(2)|.
\]
If, in addition,
\begin{equation}
    \left|
        a^2+a\beta_k+\beta_k^2
        +b_md_m+b_vd_v
    \right|
    >
    |a+\beta_k|,
    \label{eq:feedback_amplification_condition}
\end{equation}
then
\[
    |r_k(4)|>|r_k(3)|.
\]
These finite-horizon amplification conditions can hold even when
$\rho(\overline A)<1$, so asymptotic stability does not preclude transient
growth in the response.
\end{proposition}

The complete proof and spectral characterization of the frozen system are
given in Appendix~\ref{app:frozen_adamw_response}.

Proposition~\ref{prop:adamw_memory_response} makes the delay mechanism explicit.
A perturbation stored entirely in $m$ or $v$ is initially absent from the
parameter output and becomes visible only after the corresponding memory state
feeds back into the parameter update. Its subsequent magnitude depends on both
memory persistence and the return terms $b_md_m+b_vd_v$. In particular, the
second-moment feedback depends on the current momentum, gradient, and local
curvature, so it can reinforce or oppose the evolving parameter deviation.

The frozen model therefore illustrates how delayed and transiently amplified
responses can arise from AdamW memory even when the local dynamics are
asymptotically stable. The full ISO operator
\[
    c_{t+h}^{\top}\Phi_{t+h,t+1}B_t
\]
extends this mechanism to the high-dimensional, anisotropic, and time-varying
dynamics of an actual training trajectory.

\subsection{Approximation Error in Smooth and Piecewise-Smooth Networks}
\label{sec:fh_error}

The finite-horizon tangent model linearizes a nonlinear, time-varying training
trajectory. Its approximation error has two sources: the smooth Taylor
remainder within the local control region and, for piecewise-smooth networks,
the defect caused by activation-pattern changes.

Using the control-region extension $\widetilde f_s$, define
\begin{align*}
    r_s^{\mathrm{sm}}
    &:=
    \widetilde f_s(x_s^{\mathrm c}+\delta x_s)
    -
    \widetilde f_s(x_s^{\mathrm c})
    -
    A_s\delta x_s,\\
    r_s^{\mathrm{sw}}
    &:=
    f_s(x_s^{\mathrm c}+\delta x_s)
    -
    \widetilde f_s(x_s^{\mathrm c}+\delta x_s).
\end{align*}
Since $f_s$ and $\widetilde f_s$ agree at the control state, the exact
perturbation recursion is
\begin{equation}
    \delta x_{s+1}
    =
    A_s\delta x_s
    +
    r_s^{\mathrm{sm}}
    +
    r_s^{\mathrm{sw}}.
    \label{eq:piecewise_error_recursion}
\end{equation}
For smooth networks, $r_s^{\mathrm{sw}}=0$. More generally, it also vanishes
whenever the paired states remain in the same activation region.

Let the tangent prediction satisfy
\begin{equation}
    \widehat{\delta x}_{t+1}
    :=
    B_t\xi_t,
    ~~
    \widehat{\delta x}_{s+1}
    :=
    A_s\widehat{\delta x}_s.
    \label{eq:tangent_recursion}
\end{equation}

\begin{theorem}[Finite-horizon error decomposition]
\label{thm:finite_horizon_error}
Let
\(
    e_s:=\delta x_s-\widehat{\delta x}_s.
\)
Then, for every $h\geq1$,
\begin{equation}
\begin{aligned}
    e_{t+h}
    ={}&
    \Phi_{t+h,t+1}
    r_{B,t}(\xi_t)\\
    &+
    \sum_{j=t+1}^{t+h-1}
    \Phi_{t+h,j+1}
    \left(
        r_j^{\mathrm{sm}}
        +
        r_j^{\mathrm{sw}}
    \right).
\end{aligned}
\label{eq:exact_multistep_error}
\end{equation}

If $D\widetilde f_j$ is $L_j$-Lipschitz along the segment joining
$x_j^{\mathrm c}$ and $x_j^{\mathrm c}+\delta x_j$, then
\begin{equation}
\begin{aligned}
    &\|e_{t+h}\|
    \leq{}
    \|\Phi_{t+h,t+1}\|
    \|r_{B,t}(\xi_t)\|\\
    &+
    \sum_{j=t+1}^{t+h-1}
    \|\Phi_{t+h,j+1}\|
    \left(
        \frac{L_j}{2}\|\delta x_j\|^2
        +
        \|r_j^{\mathrm{sw}}\|
    \right).
\end{aligned}
\label{eq:multistep_state_error_bound}
\end{equation}

If, in addition, $F_{t+h}$ has an $L_{F,t+h}$-Lipschitz gradient along the
segment joining $\theta_{t+h}^{\mathrm c}$ and
$\theta_{t+h}^{\mathrm s}$, then
\begin{equation}
\begin{aligned}
    \bigl|
        \delta F_{t+h}
        -
        c_{t+h}^{\top}
        \Phi_{t+h,t+1}
        B_t\xi_t
    \bigr|
    &\leq{}
    \|c_{t+h}\|\,\|e_{t+h}\|\\
    &+
    \frac{L_{F,t+h}}{2}
    \|\delta\theta_{t+h}\|^2.
\end{aligned}
\label{eq:multistep_output_error_bound}
\end{equation}
\end{theorem}

Theorem~\ref{thm:finite_horizon_error} shows that the same propagators that
carry the first-order perturbation also propagate the approximation defects
introduced at each step. Large finite-horizon gain can therefore amplify both
the response of interest and the error of its tangent approximation.

\begin{corollary}[Fixed-horizon first-order accuracy]
\label{cor:fixed_horizon_accuracy}
Consider the scaled perturbation $\alpha\xi_t$ and a fixed horizon $H$.
Suppose that, in a neighborhood of the control trajectory,
$D\psi_t$ and $D\widetilde f_s$ are locally Lipschitz,
the finite-horizon propagators are uniformly bounded,
the probe objectives have locally Lipschitz gradients, and the switching
defects satisfy
\[
    \|r_s^{\mathrm{sw}}\|
    \leq
    C_s^{\mathrm{sw}}
    \|\delta x_s\|^2,
    ~~
    1\leq s-t<H.
\]
Then, for every $1\leq h\leq H$,
\[
    \delta F_{t+h}(\alpha)
    =
    \alpha
    c_{t+h}^{\top}
    \Phi_{t+h,t+1}
    B_t\xi_t
    +
    O(\alpha^2),
    ~~
    \alpha\rightarrow0.
\]
The $O(\alpha^2)$ constant may depend on the fixed horizon and control
trajectory but not on $\alpha$.
\end{corollary}

For smooth networks the switching condition holds with
$C_s^{\mathrm{sw}}=0$. For piecewise-smooth networks, a sufficient
activation-margin condition under which the aggregate switching defect is
quadratic is given in Appendix~\ref{app:fh_error}. The appendix also provides
the complete proofs and a recursive error envelope.

\section{Experiments}
\label{sec:experiments}

Our experiments address two questions.  First, does the proposed joint-state
ISO model capture the finite-horizon response mechanism
across increasingly realistic training systems?  Second, although the pathwise
ISO conditions on a realized future training sequence, does delayed influence
retain structure that is already identifiable before that future unfolds?

For mechanism validation, a control and a shock trajectory start from the same
AdamW state, differ only in the gradient applied at step $t$, and then process
the same future minibatches.  With a fixed probe objective $F$, we write
\begin{equation}
\label{eq:exp_main_response}
    d_{i,h}
    =F(\theta^{\mathrm s}_{i,t+h})-F(\theta^{\mathrm c}_{t+h}),
    ~~
    \widehat d_{i,h}
    =c_{t+h}^{\top}\Phi_{t+h,t+1}B_t\xi_i,
\end{equation}
and summarize magnitude by
$M_i=\max_{1\leq h\leq H}|d_{i,h}|$.  We evaluate trajectory fidelity, sign
agreement, and within-system Spearman correlation with $M_i$.  Candidates and
horizons are nested observations: the controlled and neural-network studies
aggregate within independent training systems, whereas the language-model
study is reported descriptively over fixed model--dataset conditions.  The
second experiment holds the present state--shock pair fixed and instead
resamples unseen future continuations.  Complete protocols, estimators, and
additional results are provided in Appendix~\ref{app:experimental_details}.

\subsection{Experiment 1: Finite-Horizon Mechanism Validation}
\label{sec:exp_main_mechanism}

\paragraph{Controlled quadratic systems.}
\label{sec:exp_main_controlled}
We begin with quadratic minibatch losses
\begin{equation}
\label{eq:exp_main_quadratic}
    \ell_s(\theta)
    =\frac12\theta^\top D_s\theta
    +\frac{1}{2r}\|U_s^\top\theta\|_2^2+q_s^\top\theta,
\end{equation}
with $d=512$, rank $r=16$, temporally correlated minibatches, and three
curvature regimes.  Four seeds with eight independently generated systems each
give 32 systems.  After 40 burn-in updates, each system uses four reference
minibatches, 16 candidate shocks, a common future of length $H=32$, and both a
standard and an anisotropic probe.  A separate rotating-readout probe is used
only in the exact-one-step-matched stress test.

We vary $\alpha\in\{1/32,1/16,1/8,1/4,1/2,1\}$.
Table~\ref{tab:exp_main_fidelity} shows that the tangent trajectory remains
accurate over the local range.  At $\alpha=1/8$, median NRMSE is $0.0483$ for
the standard probe and $0.0719$ for the anisotropic probe, with perfect median
sign agreement.  Fitting
$|d_h(\alpha)-\alpha\widehat d_h|\propto\alpha^{p_h}$ over
$\alpha\leq1/4$ gives $p_h\approx2$ throughout the horizon, with median
$R^2>0.99998$, matching the quadratic local remainder predicted by the theory.

\setlength{\textfloatsep}{4pt}
\begin{table}[t]
\centering
\caption{Representative signed-trajectory fidelity in the controlled stage of
Experiment~1.  Entries are medians over 32 independent systems after candidate
aggregation.}
\vspace{4pt}
\label{tab:exp_main_fidelity}
\small
\setlength{\tabcolsep}{3.8pt}
\begin{tabular}{@{}lcccc@{}}
\toprule
Probe & $\alpha$ & NRMSE & Rel. $M$ err. & Sign acc.\\
\midrule
Standard    & $1/32$ & 0.0117 & 0.0052 & 1.000\\
            & $1/8$  & 0.0483 & 0.0216 & 1.000\\
            & $1/4$  & 0.1007 & 0.0445 & 1.000\\
\midrule
Anisotropic & $1/32$ & 0.0173 & 0.0071 & 1.000\\
            & $1/8$  & 0.0719 & 0.0289 & 1.000\\
            & $1/4$  & 0.1473 & 0.0606 & 1.000\\
\bottomrule
\end{tabular}
\end{table}

To separate future propagation from the immediate response, we construct
shocks whose exact $|d_{i,1}|$ values are matched without using any response
at $h>1$.  The resulting within-system CV of $|d_1|$ is
$1.10\times10^{-12}$ for the standard probe and
$6.70\times10^{-12}$ for the rotating-readout probe, while the CV of future
$M$ remains $0.092$ and $0.459$.  Full ISO recovers this future ordering with
median Spearman correlations $0.993$ and $1.000$
(Table~\ref{tab:exp_main_ranking}).  Exact state interventions further
separate the parameter, first-moment, and second-moment time scales, whose
isolated responses peak near horizons $6$, $16$, and $26$, respectively.
Matched-first-displacement sweeps move the extremum later as $\beta_1$ or
$\beta_2$ increases.  These controls isolate delayed state propagation from
the immediate parameter write.

\paragraph{Nonconvex neural networks.}
\label{sec:exp_main_nonconvex}
We next apply the same paired-trajectory protocol to CIFAR-10
\citep{krizhevsky2009learning} using a $94{,}538$-parameter CNN--ReLU and an
$855{,}050$-parameter MLP--GELU.  Each
architecture contributes 16 independently trained systems, with 12 candidate
shocks per system and $H=12$ after 100 burn-in updates.  At
$\alpha=0.0625$, median trajectory NRMSE is $0.0458$ for CNN--ReLU and
$0.0377$ for MLP--GELU; at $\alpha=0.25$ it is $0.1097$ and $0.1530$.
The MLP error exponent remains $2.001$--$2.005$ across horizons, whereas the
CNN exponent decreases from $2.017$ at $h=1$ to $1.120$ at $h=12$ as
activation-pattern differences increase, consistent with the switching term
in the finite-horizon error decomposition.

At full scale, Full ISO ranks future magnitude with correlations $0.888$ and
$0.762$, compared with $0.545$ and $0.566$ for the exact one-step response.
State interventions again show an early parameter response and a later
first-moment response; increasing $\beta_1$ from $0.5$ to $0.99$ multiplies
accumulated response by $14.17$ and $11.96$ in the two architectures.  Among
the 96 exact intervention trajectories used for the signed-response
diagnostic, 57 extrema are positive and 39 are negative.

\setlength{\dbltextfloatsep}{4pt}
\begin{table*}[t]
\centering
\caption{Within-system Spearman correlation with future magnitude $M$ during
mechanism validation.  The controlled rows use exact-one-step-matched
candidates; the neural rows use natural candidates.  The one-step score is
tied in the matched controlled stress test.}
\vspace{4pt}
\label{tab:exp_main_ranking}
\small
\setlength{\tabcolsep}{8pt}
\begin{tabular}{@{}lcccc@{}}
\toprule
Setting & Full ISO & Exact 1-step & No propagation & Gradient norm\\
\midrule
Quadratic, standard probe         & 0.993 & --   & 0.354 & -0.062\\
Quadratic, rotating-readout probe & 1.000& --   & 0.806 &  0.776\\
CNN--ReLU                         & 0.888 & 0.545 & 0.755 &  0.535\\
MLP--GELU                         & 0.762 & 0.566 & 0.668 &  0.336\\
\bottomrule
\end{tabular}
\end{table*}

\paragraph{Pretrained language models.}
\label{sec:exp_main_scaling}
Finally, we evaluate Pythia-410M, Pythia-1B, and Pythia-1.4B
\citep{biderman2023pythia} on WikiText-103
\citep{merity2016pointer}, OpenWebText
\citep{gokaslan2019openwebtext}, and CodeParrot
\citep{codeparrot2022clean}.  Each of the nine model--dataset systems is
continued for 500 AdamW updates before measurement, producing nontrivial
first- and second-moment states.  Each condition then uses two reference
minibatches, eight candidate shocks, seven common-future minibatches, and
$H=8$.

At this scale we estimate the end-to-end ISO directional response numerically
using centered finite differences of the probe logits, followed by the exact
cross-entropy differential; we denote this quantity by \textsc{ISO Tangent
(FD)}.  All 72 candidates pass the adjacent-scale consistency test, with
median consistency NRMSE $0.00378$.  Trajectory NRMSE is $0.0387$, $0.0477$,
and $0.0946$ at $\alpha=0.0625$, $0.125$, and $0.25$, respectively, with
median trajectory cosine above $0.9998$ and perfect sign accuracy over this
local range.

Across model sizes, local NRMSE is $0.1090$, $0.0519$, and $0.0567$
(Table~\ref{tab:exp_main_scaling}), while trajectory cosine remains above
$0.999$. \textsc{ISO Tangent (FD)} has positive rank correlation in all nine
model--dataset conditions, with median correlations $0.714$, $0.833$, and
$0.762$ across the three model sizes. It also recovers the full-scale extremum
sign for 65 of 72 candidates. These results show that the signed
finite-horizon tangent response remains locally accurate and informative
across model scale and data domain.
\setlength{\textfloatsep}{4pt}
\begin{table}[t]
\centering
\caption{Language-model scaling results in Experiment~1. Local metrics pool
the three datasets and $\alpha\leq0.25$; rank correlations are medians over
the three fixed data-domain conditions at each model size.}
\vspace{4pt}
\label{tab:exp_main_scaling}
\small
\setlength{\tabcolsep}{3.8pt}
\begin{tabular}{@{}lccc@{}}
\toprule
Scale & Local NRMSE & Cosine & ISO-FD $\rho$\\
\midrule
0.41B & 0.1090 & 0.99918 & 0.714\\
1.0B  & 0.0519 & 0.99995 & 0.833\\
1.4B  & 0.0567 & 0.99994 & 0.762\\
\bottomrule
\end{tabular}
\end{table}

\subsection{Experiment 2: Prospective Structure Under Unknown Futures}
\label{sec:exp_main_predictability}

The pathwise operator
$c_{t+h}^{\top}\Phi_{t+h,t+1}B_t$ depends on the realized future training
sequence and is therefore not, by itself, a present-time predictor.  We ask a
more basic question: if the current AdamW state and initiating shock are held
fixed, does their finite-horizon effect remain candidate-specific when the
unseen future minibatches are resampled?

For each fixed system and candidate $i$, we draw $K=32$ independent future
continuations $\omega_k$, while sharing each continuation between its control
and shock trajectory.  Let
\begin{equation}
    M_{i,k}
    :=\max_{1\leq h\leq H}
      |d_{i,h}(\omega_k)|,
    ~~
    \mu_i:=\frac1K\sum_{k=1}^K M_{i,k}.
\end{equation}
Within each system we summarize repeated-future structure by
\begin{equation}
\label{eq:exp_main_predictability_ratio}
    \Pi_H
    :=
    \frac{\operatorname{Var}_i(\mu_i)}
    {\operatorname{Var}_i(\mu_i)
      +\mathbb E_i[\operatorname{Var}_k(M_{i,k})]}.
\end{equation}
This is a protocol-specific variance ratio, not an information-theoretic
fraction of predictable risk.  We also report the median Spearman correlation
between each branch ranking and the conditional-mean ranking.  To test whether
the structure can be extracted without observing any sampled future branch, we
compare the exact one-step response, gradient norm, parameter-write norm, a
\emph{present-frozen ISO} that repeatedly applies one current
reference-derived Jacobian with the $h=1$ probe readout frozen, and a
\emph{reference-surrogate ISO} that deterministically rolls out only the
current reference minibatches.  Neither prospective ISO score uses future
minibatches.
\setlength{\dbltextfloatsep}{6pt}
\begin{table*}[!t]
\centering
\caption{Prospective structure under 32 independently resampled future
continuations.  $\Pi_H$ is defined in
Eq.~\eqref{eq:exp_main_predictability_ratio}; branch $\rho$ is the median
single-future Spearman correlation with the conditional-mean candidate
ranking.  Present-score correlations target $\mu_i$.  All entries are medians
over independent systems.}
\vspace{4pt}
\label{tab:exp_main_predictability}
\small
\setlength{\tabcolsep}{5.2pt}
\begin{tabular}{@{}llccccc@{}}
\toprule
System & Candidates & $\Pi_H$ & Branch $\rho$ & 1-step $\rho$
& Frozen ISO $\rho$ & Ref.-surrogate $\rho$\\
\midrule
Quadratic, standard & Natural & 0.801 & 0.919 & 0.801 & 0.659 & 0.372\\
                    & Matched & 0.730 & 0.909 & --   & -0.066 & -0.021\\
MLP--GELU           & Natural & 0.771 & 0.862 & 0.661 & 0.734 & 0.262\\
                    & Matched & 0.921 & 0.955 & --   & 0.941 & 0.752\\
CNN--ReLU           & Natural & 0.716 & 0.855 & 0.619 & 0.752 & 0.601\\
                    & Matched & 0.700 & 0.872 & --   & 0.755 & 0.811\\
\bottomrule
\end{tabular}
\end{table*}

\paragraph{Controlled repeated futures.}
In the standard quadratic regime, natural candidates give
$\Pi_{32}=0.801$, but their exact one-step response already correlates
$0.801$ with $\mu_i$.  We therefore repeat the analysis after exact one-step
matching.  The matched $|d_1|$ has median within-system CV
$1.07\times10^{-12}$, yet $\Pi_{32}$ remains $0.730$ and the median
single-branch ranking correlation with $\mu_i$ is $0.909$.  Thus the delayed
response retains substantial candidate-specific structure after immediate
magnitude is removed.  However, this structure is not recovered by the
simplest present-time compressions in the controlled system: correlations with
$\mu_i$ are $0.049$ for gradient norm, $-0.057$ for parameter-write norm, and
$-0.066$ for present-frozen ISO.  The anisotropic-probe results show the same
qualitative separation and are reported in Appendix~\ref{app:exp2_controlled_predictability}.

\paragraph{Neural-network repeated futures.}
The same construction is applied to the fixed post-burn-in CIFAR-10 systems.
Exact one-step matching succeeds for every candidate, with median CV
$3.57\times10^{-13}$ for MLP--GELU and $1.36\times10^{-12}$ for CNN--ReLU.
After matching, prospective structure remains strong:
$\Pi_{12}=0.921$ for MLP--GELU and $0.700$ for CNN--ReLU.  Unlike the
controlled setting, present-time ISO scores now recover much of the
conditional-mean ordering.  Present-frozen ISO reaches median correlations
$0.941$ and $0.755$, compared with gradient-norm correlations $0.066$ and
$0.500$ for MLP--GELU and CNN--ReLU, respectively.  The
reference-surrogate ISO reaches $0.752$ and $0.811$.  The stronger frozen-ISO
agreement in the smooth MLP is consistent with the greater local tangent
coherence observed in Experiment~1.

\begin{remark}[Why quadratic systems perform worse]
The quadratic
system is globally smooth, yet its frozen ISO performs poorly after one-step
matching. A more plausible factor is how well the current local dynamics
represent the future time-varying propagation. Longer horizons, transition and
readout drift, and activation switching can all reduce this coherence.
\end{remark}

Taken together, Experiment~1 establishes the pathwise finite-horizon mechanism,
whereas Experiment~2 shows that its delayed effects are not created entirely
by the subsequently realized minibatches: substantial candidate-specific
structure can already be present at the perturbation time.  Whether that
structure admits an accurate present-time representation is regime-dependent,
as illustrated by the contrast between the controlled, MLP--GELU, and
CNN--ReLU results.  Characterizing the conditions for such prospective
identifiability is distinct from the pathwise mechanism studied here.

\section{Conclusion and Discussion}

We studied delayed minibatch influence in AdamW through a finite-horizon
input--state--output formulation that tracks how perturbations enter optimizer
state, propagate through future updates, and appear in later losses. The
resulting tangent model is supported across controlled systems, neural
networks, and pretrained language models, while repeated-future experiments
show that delayed influence can also contain prospective structure. The main
limitations are that the exact ISO is pathwise and its present-time
approximation may degrade under nonlinear dynamics, activation switching, and
future dynamical drift. Future work should characterize when such prospective
influence is identifiable from the current optimizer state, ideally through
necessary and sufficient conditions, and determine whether this can support
reliable training-time warning or control.

\section*{Acknowledgments}

We gratefully acknowledge Hongqian Huang for providing the computational
resources used in this work.

\bibliography{example_paper}
\bibliographystyle{icml2025}

\newpage
\appendix


\section{Detailed Derivation of the Minibatch Shock Write-In}
\label{app:writein}

This appendix derives the input operator $B_t$, its nonlinear remainder, and
the second-order gradient-to-parameter map used in the one-step output
expansion.

\subsection{AdamW Update as a State Transition}
\label{app:adamw_transition}

For a fixed step $s$, define
\[
    \rho_{1,s}:=1-\beta_1^s,
    ~~
    \rho_{2,s}:=1-\beta_2^s.
\]
Following AdamW \citep{loshchilov2019decoupled}, given the pre-update state
\[
x_s=
\begin{bmatrix}
\theta_s\\
m_{s-1}\\
v_{s-1}
\end{bmatrix},
\]
and an input gradient $g\in\mathbb R^d$, the moment states after the update are
\begin{align}
    m_s(g)
    &:=
    \beta_1m_{s-1}
    +(1-\beta_1)g,
    \label{eq:app_m_of_g}\\
    v_s(g)
    &:=
    \beta_2v_{s-1}
    +(1-\beta_2)g^{\odot2}.
    \label{eq:app_v_of_g}
\end{align}
The bias-corrected moments are
\[
    \widehat m_s(g)
    :=
    \frac{m_s(g)}{\rho_{1,s}},
    ~~
    \widehat v_s(g)
    :=
    \frac{v_s(g)}{\rho_{2,s}},
\]
and the adaptive parameter displacement is
\[
    q_s(m,v)
    :=
    -\eta_s
    \left(
        \frac{m}{\rho_{1,s}}
    \right)
    \oslash
    \left(
        \sqrt{\frac{v}{\rho_{2,s}}}
        +\epsilon\mathbf 1
    \right).
\]
Define the composite gradient-to-parameter map
\[
    \psi_s(g)
    :=
    q_s\bigl(m_s(g),v_s(g)\bigr).
\]
At the shock step,
\[
    \theta_{t+1}(g)
    =
    D_t\theta_t+\psi_t(g).
\]
Because the control and shock trajectories share the same pre-update
parameter $\theta_t$, the term $D_t\theta_t$ cancels from their difference.

All derivatives below are evaluated along the control trajectory. Whenever a
derivative with respect to the raw second-moment state is used, we assume
\[
    \widehat v_{s,i}^{\mathrm c}>0
\]
for the corresponding coordinates.

\subsection{Moment-State Sensitivities}
\label{app:moment_sensitivities}

Write
\[
    d_s
    :=
    \sqrt{\widehat v_s^{\mathrm c}}
    +\epsilon\mathbf 1.
\]
Since the adaptive map is coordinate-separable, its derivatives with respect
to $m$ and $v$ are diagonal.

For coordinate $i$,
\[
    q_{s,i}(m_i,v_i)
    =
    -\eta_s
    \frac{m_i/\rho_{1,s}}
    {\sqrt{v_i/\rho_{2,s}}+\epsilon}.
\]
Differentiating with respect to $m_i$ gives
\[
    \frac{\partial q_{s,i}}{\partial m_i}
    =
    -\frac{\eta_s}
    {\rho_{1,s}d_{s,i}}.
\]
Hence
\begin{equation}
    M_s
    =
    \frac{\partial q_s}{\partial m_s}
    =
    -\frac{\eta_s}{\rho_{1,s}}
    \operatorname{Diag}(d_s^{-1}).
    \label{eq:app_M}
\end{equation}

For the derivative with respect to $v_i$, define
\[
    r_{s,i}
    :=
    \widehat v_{s,i}^{\mathrm c}
    =
    \frac{v_{s,i}^{\mathrm c}}{\rho_{2,s}}.
\]
Using
\[
    \frac{\partial \sqrt{v_i/\rho_{2,s}}}{\partial v_i}
    =
    \frac{1}
    {2\rho_{2,s}\sqrt{r_{s,i}}},
\]
we obtain
\[
\begin{aligned}
    \frac{\partial q_{s,i}}{\partial v_i}
    &=
    \eta_s
    \frac{m_{s,i}^{\mathrm c}/\rho_{1,s}}
    {2\rho_{2,s}
     \sqrt{r_{s,i}}
     \left(\sqrt{r_{s,i}}+\epsilon\right)^2}\\
    &=
    \eta_s
    \frac{\widehat m_{s,i}^{\mathrm c}}
    {2\rho_{2,s}
     \sqrt{\widehat v_{s,i}^{\mathrm c}}
     d_{s,i}^2}.
\end{aligned}
\]
Therefore
\begin{equation}
    V_s
    =
    \frac{\partial q_s}{\partial v_s}
    =
    \eta_s
    \operatorname{Diag}
    \left(
        \frac{
            \widehat m_s^{\mathrm c}
        }{
            2\rho_{2,s}
            \sqrt{\widehat v_s^{\mathrm c}}
            \odot d_s^{\odot2}
        }
    \right).
    \label{eq:app_V}
\end{equation}

Equation~\eqref{eq:app_M} shows that $M_s$ is negative diagonal. The sign of
the $i$th diagonal entry of $V_s$ is the sign of
$\widehat m_{s,i}^{\mathrm c}$.

\subsection{Gradient-to-Parameter Jacobian}
\label{app:gradient_parameter_jacobian}

The derivatives of the moment maps in
Equations~\eqref{eq:app_m_of_g}--\eqref{eq:app_v_of_g} are
\begin{align*}
    Dm_s(g)[u]
    &=
    (1-\beta_1)u,\\
    Dv_s(g)[u]
    &=
    2(1-\beta_2)
    \operatorname{Diag}(g)u.
\end{align*}
Applying the chain rule to
$\psi_s(g)=q_s(m_s(g),v_s(g))$ gives
\[
\begin{aligned}
    D\psi_s(g_s^{\mathrm c})[u]
    &=
    M_sDm_s(g_s^{\mathrm c})[u]
    +
    V_sDv_s(g_s^{\mathrm c})[u]\\
    &=
    \Bigl[
        (1-\beta_1)M_s
        +
        2(1-\beta_2)
        V_s\operatorname{Diag}(g_s^{\mathrm c})
    \Bigr]u.
\end{aligned}
\]
Thus
\begin{equation}
    \Gamma_s
    :=
    D\psi_s(g_s^{\mathrm c})
    =
    (1-\beta_1)M_s
    +
    2(1-\beta_2)
    V_s\operatorname{Diag}(g_s^{\mathrm c}).
    \label{eq:app_Gamma}
\end{equation}
The two terms correspond to the differential paths
\[
    g\rightarrow m\rightarrow\theta,
    ~~
    g\rightarrow v\rightarrow\theta.
\]

\subsection{Exact Shock-Step State Difference}
\label{app:exact_writein}

At the shock step,
\[
    g_t^{\mathrm s}
    =
    g_t^{\mathrm c}+\xi_t.
\]
Subtracting the first-moment updates gives
\begin{equation}
\begin{aligned}
    \delta m_t
    &=
    m_t^{\mathrm s}-m_t^{\mathrm c}\\
    &=
    (1-\beta_1)
    \left(
        g_t^{\mathrm s}-g_t^{\mathrm c}
    \right)\\
    &=
    (1-\beta_1)\xi_t.
\end{aligned}
\label{eq:app_exact_delta_m}
\end{equation}

For the second moment,
\begin{equation}
\begin{aligned}
    \delta v_t
    &=
    (1-\beta_2)
    \left[
        (g_t^{\mathrm c}+\xi_t)^{\odot2}
        -
        (g_t^{\mathrm c})^{\odot2}
    \right]\\
    &=
    2(1-\beta_2)
    g_t^{\mathrm c}\odot\xi_t
    +
    (1-\beta_2)\xi_t^{\odot2}\\
    &=
    2(1-\beta_2)
    \operatorname{Diag}(g_t^{\mathrm c})\xi_t
    +
    (1-\beta_2)\xi_t^{\odot2}.
\end{aligned}
\label{eq:app_exact_delta_v}
\end{equation}

The parameter difference is
\begin{equation}
\begin{aligned}
    \delta\theta_{t+1}
    &=
    \psi_t(g_t^{\mathrm c}+\xi_t)
    -
    \psi_t(g_t^{\mathrm c})\\
    &=
    \Gamma_t\xi_t
    +
    r_{\theta,t}(\xi_t),
\end{aligned}
\label{eq:app_delta_theta}
\end{equation}
where
\[
    r_{\theta,t}(\xi)
    :=
    \psi_t(g_t^{\mathrm c}+\xi)
    -
    \psi_t(g_t^{\mathrm c})
    -
    \Gamma_t\xi.
\]

Stacking
Equations~\eqref{eq:app_exact_delta_m},
\eqref{eq:app_exact_delta_v}, and
\eqref{eq:app_delta_theta} gives
\[
    \delta x_{t+1}
    =
    B_t\xi_t+r_{B,t}(\xi_t),
\]
where
\[
    B_t
    :=
    \begin{bmatrix}
        \Gamma_t\\[1mm]
        (1-\beta_1)I\\[1mm]
        2(1-\beta_2)
        \operatorname{Diag}(g_t^{\mathrm c})
    \end{bmatrix}
\]
and
\[
    r_{B,t}(\xi)
    :=
    \begin{bmatrix}
        r_{\theta,t}(\xi)\\[1mm]
        0\\[1mm]
        (1-\beta_2)\xi^{\odot2}
    \end{bmatrix}.
\]

If $D\psi_t$ is $\kappa_{\theta,t}$-Lipschitz in a neighborhood containing
the segment
\[
    \left\{
        g_t^{\mathrm c}+\tau\xi_t:
        0\leq\tau\leq1
    \right\},
\]
Taylor's theorem gives
\[
    \|r_{\theta,t}(\xi_t)\|
    \leq
    \frac{\kappa_{\theta,t}}{2}
    \|\xi_t\|^2.
\]
Moreover,
\[
    \|\xi^{\odot2}\|_2
    =
    \left(
        \sum_i\xi_i^4
    \right)^{1/2}
    \leq
    \sum_i\xi_i^2
    =
    \|\xi\|_2^2.
\]
Consequently,
\begin{equation}
    \|r_{B,t}(\xi_t)\|_2
    \leq
    \left(
        \frac{\kappa_{\theta,t}}{2}
        +
        1-\beta_2
    \right)
    \|\xi_t\|_2^2.
    \label{eq:app_writein_quadratic_bound}
\end{equation}
For a scaled perturbation $\alpha\xi_t$,
Equation~\eqref{eq:app_writein_quadratic_bound} implies
\[
    \|r_{B,t}(\alpha\xi_t)\|
    =
    O(\alpha^2).
\]

\subsection{Second Derivative of the Gradient-to-Parameter Map}
\label{app:second_order_writein}

For completeness, we derive
\[
    \mathcal Q_t
    :=
    D^2\psi_t(g_t^{\mathrm c}),
\]
which is used in the one-step second-order output expansion.

Let
\[
    a_1:=1-\beta_1,
    ~~
    a_2:=1-\beta_2.
\]
The moment maps satisfy
\begin{align*}
    Dm_t(g)[u]
    &=
    a_1u,\\
    D^2m_t(g)[u,w]
    &=
    0,\\
    Dv_t(g)[u]
    &=
    2a_2\operatorname{Diag}(g)u,\\
    D^2v_t(g)[u,w]
    &=
    2a_2(u\odot w).
\end{align*}

Because $q_t$ is coordinate-separable, $\mathcal Q_t$ is also
coordinate-separable. Define
\[
    s_{t,i}
    :=
    \sqrt{\widehat v_{t,i}^{\mathrm c}},
    ~~
    d_{t,i}
    :=
    s_{t,i}+\epsilon.
\]
The nonzero second derivatives of $q_{t,i}$ are
\begin{align*}
    q_{mv,t,i}
    &:=
    \frac{\partial^2q_{t,i}}
    {\partial m_i\partial v_i}
    =
    \frac{\eta_t}
    {2\rho_{1,t}\rho_{2,t}
     s_{t,i}d_{t,i}^2},\\
    q_{vv,t,i}
    &:=
    \frac{\partial^2q_{t,i}}
    {\partial v_i^2}
    =
    -\frac{
        \eta_t\widehat m_{t,i}^{\mathrm c}
        (3s_{t,i}+\epsilon)
    }{
        4\rho_{2,t}^2
        s_{t,i}^3
        d_{t,i}^3
    }.
\end{align*}
Also,
\[
    q_{v,t,i}
    =
    [V_t]_{ii}.
\]

The second-order chain rule gives
\[
\begin{aligned}
    \mathcal Q_t[u,w]
    ={}&
    D_{mv}^2q_t
    \left[
        Dm_t[u],Dv_t[w]
    \right]\\
    &+
    D_{vm}^2q_t
    \left[
        Dv_t[u],Dm_t[w]
    \right]\\
    &+
    D_{vv}^2q_t
    \left[
        Dv_t[u],Dv_t[w]
    \right]\\
    &+
    V_tD^2v_t[u,w].
\end{aligned}
\]
Coordinatewise,
\[
    [\mathcal Q_t[u,w]]_i
    =
    \chi_{t,i}u_iw_i,
\]
where
\[
\begin{aligned}
    \chi_{t,i}
    :={}&
    4a_1a_2
    g_{t,i}^{\mathrm c}
    q_{mv,t,i}\\
    &+
    4a_2^2
    (g_{t,i}^{\mathrm c})^2
    q_{vv,t,i}
    +
    2a_2[V_t]_{ii}.
\end{aligned}
\]
Therefore, for a scaled perturbation $\alpha\xi_t$,
\begin{equation}
    \delta\theta_{t+1}(\alpha)
    =
    \alpha\Gamma_t\xi_t
    +
    \frac{\alpha^2}{2}
    \mathcal Q_t[\xi_t,\xi_t]
    +
    o(\alpha^2).
    \label{eq:app_second_order_theta}
\end{equation}


\section{Proof of the Finite-Horizon Directional Response}
\label{app:fh_response}

This appendix derives the joint AdamW transition Jacobian, proves
Theorem~\ref{thm:finite_horizon_response}, and records several properties of
the resulting input--output operator.

\subsection{Control-Region State Transition}
\label{app:control_region_transition}

Let $f_s$ denote the exact AdamW transition under future minibatch
$\mathcal B_s$:
\[
    x_{s+1}=f_s(x_s).
\]
For a smooth network, define
$\widetilde f_s:=f_s$.
For a piecewise-smooth network, let $\widetilde f_s$ denote the smooth
extension associated with the activation pattern of the control trajectory at
$x_s^{\mathrm c}$.

The control-region tangent matrix is
\[
    A_s
    :=
    D\widetilde f_s(x_s^{\mathrm c}).
\]
All quantities in the following block derivation are evaluated at the control
state and its corresponding future minibatch.

Let $\widetilde g_s(\theta)$ denote the gradient map induced by the same
control-region smooth extension, and define
\[
    H_s^{\mathrm{tr}}
    :=
    D\widetilde g_s(\theta_s^{\mathrm c}).
\]
When the loss is twice differentiable in the control region,
\[
    H_s^{\mathrm{tr}}
    =
    \nabla_\theta^2
    \ell(\theta_s^{\mathrm c};\mathcal B_s).
\]

\subsection{Blockwise Derivation of the Joint Jacobian}
\label{app:block_jacobian}

For a generic state
\[
    x
    =
    [\theta,m_-,v_-]^\top,
\]
the control-region transition has components
\begin{align*}
    m^+
    &=
    \beta_1m_-
    +(1-\beta_1)\widetilde g_s(\theta),\\
    v^+
    &=
    \beta_2v_-
    +(1-\beta_2)
    \widetilde g_s(\theta)^{\odot2},\\
    \theta^+
    &=
    D_s\theta+q_s(m^+,v^+).
\end{align*}

The derivatives of the first-moment update are
\begin{align*}
    \frac{\partial m^+}{\partial\theta}
    &=
    (1-\beta_1)H_s^{\mathrm{tr}},\\
    \frac{\partial m^+}{\partial m_-}
    &=
    \beta_1I,\\
    \frac{\partial m^+}{\partial v_-}
    &=
    0.
\end{align*}

For the second moment,
\[
    D
    \left[
        \widetilde g_s(\theta)^{\odot2}
    \right]
    =
    2\operatorname{Diag}
    \left(
        \widetilde g_s(\theta)
    \right)
    D\widetilde g_s(\theta),
\]
so
\begin{align*}
    \frac{\partial v^+}{\partial\theta}
    &=
    2(1-\beta_2)
    \operatorname{Diag}(g_s^{\mathrm c})
    H_s^{\mathrm{tr}}
    :=
    S_s,\\
    \frac{\partial v^+}{\partial m_-}
    &=
    0,\\
    \frac{\partial v^+}{\partial v_-}
    &=
    \beta_2I.
\end{align*}

For the parameter update,
\[
\begin{aligned}
    \frac{\partial\theta^+}{\partial\theta}
    ={}&
    D_s
    +
    M_s
    \frac{\partial m^+}{\partial\theta}
    +
    V_s
    \frac{\partial v^+}{\partial\theta}\\
    ={}&
    D_s
    +
    (1-\beta_1)M_sH_s^{\mathrm{tr}}\\
    &+
    2(1-\beta_2)
    V_s\operatorname{Diag}(g_s^{\mathrm c})
    H_s^{\mathrm{tr}}\\
    ={}&
    D_s+\Gamma_sH_s^{\mathrm{tr}}.
\end{aligned}
\]
Similarly,
\begin{align*}
    \frac{\partial\theta^+}{\partial m_-}
    &=
    \beta_1M_s,\\
    \frac{\partial\theta^+}{\partial v_-}
    &=
    \beta_2V_s.
\end{align*}

Combining the nine blocks gives
\begin{equation}
    A_s
    =
    \begin{bmatrix}
        D_s+\Gamma_sH_s^{\mathrm{tr}}
        &
        \beta_1M_s
        &
        \beta_2V_s
        \\[1mm]
        (1-\beta_1)H_s^{\mathrm{tr}}
        &
        \beta_1I
        &
        0
        \\[1mm]
        S_s
        &
        0
        &
        \beta_2I
    \end{bmatrix}.
    \label{eq:app_joint_jacobian}
\end{equation}

For later use, recall the general propagator convention
\[
    \Phi_{a,b}
    :=
    \begin{cases}
        I,
        &a=b,\\[1mm]
        A_{a-1}A_{a-2}\cdots A_b,
        &a>b.
    \end{cases}
\]

\subsection{Proof of Theorem~\ref{thm:finite_horizon_response}}
\label{app:proof_fh_response}

Consider
\[
    g_t^{\mathrm s}(\alpha)
    =
    g_t^{\mathrm c}
    +
    \alpha\xi_t.
\]
At $\alpha=0$, the shock and control trajectories coincide.

By assumption, the control states lie in the interior of the smooth regions
used to define the transitions
$\widetilde f_{t+1},\ldots,\widetilde f_{t+h-1}$.
For a fixed horizon, continuity of the trajectory implies that there exists
$\alpha_0>0$ such that, for sufficiently small $|\alpha|<\alpha_0$, the
perturbed trajectory follows the same sequence of local smooth extensions.
The differentiability assumptions on the AdamW transition, including the
required positivity of the second-moment coordinates, ensure that the
Jacobians used below are well defined.

Define
\[
    \dot x_s
    :=
    \left.
    \frac{\mathrm d}{\mathrm d\alpha}
    \delta x_s(\alpha)
    \right|_{\alpha=0}.
\]
By Appendix~\ref{app:writein},
\[
    \dot x_{t+1}
    =
    B_t\xi_t.
\]

For every future step $s\geq t+1$,
\[
    x_{s+1}^{\mathrm s}(\alpha)
    =
    \widetilde f_s
    \left(
        x_s^{\mathrm s}(\alpha)
    \right),
\]
while
\[
    x_{s+1}^{\mathrm c}
    =
    \widetilde f_s
    \left(
        x_s^{\mathrm c}
    \right).
\]
Differentiating at $\alpha=0$ yields
\begin{equation}
    \dot x_{s+1}
    =
    D\widetilde f_s(x_s^{\mathrm c})
    \dot x_s
    =
    A_s\dot x_s.
    \label{eq:app_tangent_recursion}
\end{equation}
Repeated application gives
\[
\begin{aligned}
    \dot x_{t+h}
    &=
    A_{t+h-1}
    A_{t+h-2}
    \cdots
    A_{t+1}
    B_t\xi_t\\
    &=
    \Phi_{t+h,t+1}
    B_t\xi_t.
\end{aligned}
\]
For $h=1$,
$\Phi_{t+1,t+1}=I$.

The probe function at horizon $t+h$ is shared by the paired trajectories, so
\[
    \delta F_{t+h}(\alpha)
    =
    F_{t+h}
    \left(
        \theta_{t+h}^{\mathrm s}(\alpha)
    \right)
    -
    F_{t+h}
    \left(
        \theta_{t+h}^{\mathrm c}
    \right).
\]
Differentiating at $\alpha=0$ gives
\[
\begin{aligned}
    \left.
    \frac{\mathrm d}{\mathrm d\alpha}
    \delta F_{t+h}(\alpha)
    \right|_{\alpha=0}
    &=
    \nabla F_{t+h}
    \left(
        \theta_{t+h}^{\mathrm c}
    \right)^\top
    \dot\theta_{t+h}\\
    &=
    c_{t+h}^{\top}\dot x_{t+h}\\
    &=
    c_{t+h}^{\top}
    \Phi_{t+h,t+1}
    B_t\xi_t.
\end{aligned}
\]
This proves both directional identities in
Theorem~\ref{thm:finite_horizon_response}. The first-order expansion
\[
    \delta F_{t+h}(\alpha)
    =
    \alpha
    c_{t+h}^{\top}
    \Phi_{t+h,t+1}
    B_t\xi_t
    +
    o(\alpha)
\]
follows directly from differentiability at $\alpha=0$.

\subsection{Complete One-Step Second-Order Output Expansion}
\label{app:one_step_output}

We now recover the one-step second-order geometry that is omitted from the
main text.

From Equation~\eqref{eq:app_second_order_theta},
\[
    \delta\theta_{t+1}(\alpha)
    =
    \alpha p_t
    +
    \frac{\alpha^2}{2}u_t
    +
    o(\alpha^2),
\]
where
\[
    p_t
    :=
    \Gamma_t\xi_t,
    ~~
    u_t
    :=
    \mathcal Q_t[\xi_t,\xi_t].
\]

Assume that $F_{t+1}$ is twice differentiable in the relevant local region and
define
\[
    H_{t+1}^{F}
    :=
    \nabla^2F_{t+1}
    \left(
        \theta_{t+1}^{\mathrm c}
    \right).
\]
Taylor expansion around $\theta_{t+1}^{\mathrm c}$ gives
\[
\begin{aligned}
    \delta F_{t+1}(\alpha)
    ={}&
    \nabla F_{t+1}
    \left(
        \theta_{t+1}^{\mathrm c}
    \right)^\top
    \delta\theta_{t+1}(\alpha)\\
    &+
    \frac12
    \delta\theta_{t+1}(\alpha)^\top
    H_{t+1}^{F}
    \delta\theta_{t+1}(\alpha)\\
    &+
    o\left(
        \|\delta\theta_{t+1}(\alpha)\|^2
    \right).
\end{aligned}
\]
The linear output contribution is
\[
\begin{aligned}
    \nabla F_{t+1}^\top
    \delta\theta_{t+1}(\alpha)
    ={}&
    \alpha
    \nabla F_{t+1}^\top p_t\\
    &+
    \frac{\alpha^2}{2}
    \nabla F_{t+1}^\top u_t
    +
    o(\alpha^2),
\end{aligned}
\]
where the gradients are evaluated at
$\theta_{t+1}^{\mathrm c}$.
The quadratic output contribution satisfies
\[
    \frac12
    \delta\theta_{t+1}(\alpha)^\top
    H_{t+1}^{F}
    \delta\theta_{t+1}(\alpha)
    =
    \frac{\alpha^2}{2}
    p_t^\top H_{t+1}^{F}p_t
    +
    o(\alpha^2).
\]
Therefore
\begin{equation}
\begin{aligned}
    \delta F_{t+1}(\alpha)
    ={}&
    \alpha
    \nabla F_{t+1}
    \left(
        \theta_{t+1}^{\mathrm c}
    \right)^\top
    \Gamma_t\xi_t\\
    &+
    \frac{\alpha^2}{2}
    \Bigl[
        (\Gamma_t\xi_t)^\top
        H_{t+1}^{F}
        (\Gamma_t\xi_t)\\
    &\hspace{16mm}
        +
        \nabla F_{t+1}
        \left(
            \theta_{t+1}^{\mathrm c}
        \right)^\top
        \mathcal Q_t[\xi_t,\xi_t]
    \Bigr]\\
    &+
    o(\alpha^2).
\end{aligned}
\label{eq:app_complete_one_step}
\end{equation}

The first second-order term in
Equation~\eqref{eq:app_complete_one_step} is the probe-curvature contribution
induced by the first-order parameter displacement. Equivalently, it is
generated by the effective curvature operator
\[
    \Gamma_t^\top
    H_{t+1}^{F}
    \Gamma_t.
\]
The second term is the output effect of the nonlinear AdamW write-in itself.
The training-batch Hessian $H_s^{\mathrm{tr}}$ governs the subsequent state
propagation, whereas $H_s^F$ describes curvature of the probe output. Neither
curvature term by itself determines the sign of the response.

\subsection{Coordinate Invariance of the Input--Output Operator}
\label{app:coordinate_invariance}

Let
\[
    \widetilde x_s
    :=
    T_sx_s
\]
for invertible matrices $T_s$. The transformed state transition is
\[
    \widetilde A_s
    :=
    T_{s+1}A_sT_s^{-1}.
\]
Hence
\[
\begin{aligned}
    \widetilde\Phi_{t+h,t+1}
    &=
    \widetilde A_{t+h-1}
    \cdots
    \widetilde A_{t+1}\\
    &=
    T_{t+h}
    A_{t+h-1}
    T_{t+h-1}^{-1}
    \cdots
    T_{t+2}
    A_{t+1}
    T_{t+1}^{-1}\\
    &=
    T_{t+h}
    \Phi_{t+h,t+1}
    T_{t+1}^{-1}.
\end{aligned}
\]
The input and output maps transform as
\[
    \widetilde B_t
    =
    T_{t+1}B_t,
    ~~
    \widetilde c_{t+h}^{\top}
    =
    c_{t+h}^{\top}T_{t+h}^{-1}.
\]
Therefore
\[
\begin{aligned}
    \widetilde c_{t+h}^{\top}
    \widetilde\Phi_{t+h,t+1}
    \widetilde B_t
    &=
    c_{t+h}^{\top}
    T_{t+h}^{-1}
    T_{t+h}
    \Phi_{t+h,t+1}
    T_{t+1}^{-1}
    T_{t+1}
    B_t\\
    &=
    c_{t+h}^{\top}
    \Phi_{t+h,t+1}
    B_t.
\end{aligned}
\]
Thus the finite-horizon input--output operator is invariant under invertible
state reparameterization. Internal state-gain quantities can depend on the
relative scaling chosen for the parameter and moment blocks, whereas the
signed scalar input--output response does not.

\subsection{Additional Finite-Horizon Response Summaries}
\label{app:response_summaries}

The main text uses the maximum response magnitude, its timing, its sign, and
the largest positive excursion as the primary finite-horizon summaries. We
record additional cumulative and direction-specific quantities here.

The accumulated absolute response is
\[
    \operatorname{ARE}_{t,H}
    :=
    \sum_{h=1}^{H}
    |\delta F_{t+h}|.
\]
While $M_{t,H}$ measures the largest deviation over the horizon,
$\operatorname{ARE}_{t,H}$ measures the total magnitude accumulated along the
response trajectory.

The largest negative excursion is
\[
    P_{t,H}^{-}
    :=
    \max_{1\leq h\leq H}
    [-\delta F_{t+h}]_+.
\]
Together,
$P_{t,H}^{+}$ and $P_{t,H}^{-}$ distinguish the largest positive and negative
deviations from the control trajectory.

Their accumulated counterparts are
\[
    \operatorname{AEL}_{t,H}^{+}
    :=
    \sum_{h=1}^{H}
    [\delta F_{t+h}]_+,
    ~~
    \operatorname{AEL}_{t,H}^{-}
    :=
    \sum_{h=1}^{H}
    [-\delta F_{t+h}]_+.
\]
The corresponding tangent quantities are obtained by replacing
$\delta F_{t+h}$ with
$\mathcal G_{t,h}\xi_t$. For example,
\[
    \widehat{\operatorname{ARE}}_{t,H}
    :=
    \sum_{h=1}^{H}
    |\mathcal G_{t,h}\xi_t|,
\]
and
\[
    \widehat P_{t,H}^{-}
    :=
    \max_{1\leq h\leq H}
    [-\mathcal G_{t,h}\xi_t]_+.
\]
These quantities are secondary summaries of the same signed finite-horizon
response rather than separate dynamical objects.


\section{Frozen AdamW Memory-Channel Analysis}
\label{app:frozen_adamw_response}

This appendix proves
Proposition~\ref{prop:adamw_memory_response}
and gives spectral, transfer-function, and feedback-loop characterizations of
the frozen three-state model.

\subsection{Short-Horizon Responses}
\label{app:short_horizon_responses}

Consider
\[
    \overline A
    =
    \begin{bmatrix}
        a & b_m & b_v\\
        d_m & \beta_1 & 0\\
        d_v & 0 & \beta_2
    \end{bmatrix},
\]
with normalized parameter readout
\[
    e_\theta
    :=
    \begin{bmatrix}
        1\\0\\0
    \end{bmatrix},
\]
and memory-channel basis vectors
\[
    e_m
    :=
    \begin{bmatrix}
        0\\1\\0
    \end{bmatrix},
    ~~
    e_v
    :=
    \begin{bmatrix}
        0\\0\\1
    \end{bmatrix}.
\]
For $k\in\{m,v\}$, define
\[
    r_k(h)
    :=
    e_\theta^\top
    \overline A^{h-1}
    e_k.
\]

If the scalar probe sensitivity at the frozen operating point is
$\gamma_F\neq0$, then the corresponding first-order probe-loss response is
$\gamma_Fr_k(h)$. Thus, $r_k(h)$ isolates the timing and amplification produced
by the memory-to-parameter dynamics, while the probe readout supplies the final
output scaling and sign.

At $h=1$,
\[
    r_m(1)=r_v(1)=0.
\]
At $h=2$,
\[
    \overline A e_m
    =
    \begin{bmatrix}
        b_m\\
        \beta_1\\
        0
    \end{bmatrix},
    ~~
    \overline A e_v
    =
    \begin{bmatrix}
        b_v\\
        0\\
        \beta_2
    \end{bmatrix},
\]
so
\[
    r_m(2)=b_m,
    ~~
    r_v(2)=b_v.
\]

Applying $\overline A$ again,
\[
    \overline A^2e_m
    =
    \begin{bmatrix}
        b_m(a+\beta_1)\\
        b_md_m+\beta_1^2\\
        b_md_v
    \end{bmatrix},
\]
and
\[
    \overline A^2e_v
    =
    \begin{bmatrix}
        b_v(a+\beta_2)\\
        b_vd_m\\
        b_vd_v+\beta_2^2
    \end{bmatrix}.
\]
Hence
\[
    r_m(3)
    =
    b_m(a+\beta_1),
    ~~
    r_v(3)
    =
    b_v(a+\beta_2).
\]

A third multiplication gives
\[
\begin{aligned}
    r_m(4)
    &=
    a\,b_m(a+\beta_1)
    +
    b_m(b_md_m+\beta_1^2)
    +
    b_vb_md_v\\
    &=
    b_m
    \left(
        a^2+a\beta_1+\beta_1^2
        +b_md_m+b_vd_v
    \right),
\end{aligned}
\]
and
\[
\begin{aligned}
    r_v(4)
    &=
    a\,b_v(a+\beta_2)
    +
    b_mb_vd_m
    +
    b_v(b_vd_v+\beta_2^2)\\
    &=
    b_v
    \left(
        a^2+a\beta_2+\beta_2^2
        +b_md_m+b_vd_v
    \right).
\end{aligned}
\]
Therefore,
\[
    r_k(4)
    =
    b_k
    \left(
        a^2+a\beta_k+\beta_k^2
        +b_md_m+b_vd_v
    \right).
\]

If $b_k\neq0$, then
\[
    |r_k(3)|>|r_k(2)|
\]
is equivalent to
\[
    |a+\beta_k|>1.
\]
Likewise,
\[
    |r_k(4)|>|r_k(3)|
\]
holds whenever
\[
    \left|
        a^2+a\beta_k+\beta_k^2
        +b_md_m+b_vd_v
    \right|
    >
    |a+\beta_k|.
\]

These are finite-horizon algebraic conditions and do not require asymptotic
stability. When they hold together with
\[
    \rho(\overline A)<1,
\]
the response grows over the corresponding short horizon even though
\[
    \overline A^h\rightarrow0
    ~~
    \text{as }h\rightarrow\infty.
\]
This is the transient amplification regime described in the main text.

\subsection{Characteristic Polynomial}
\label{app:frozen_characteristic}

The characteristic polynomial is
\begin{equation}
\begin{aligned}
    p(\lambda)
    &:=
    \det(\lambda I-\overline A)\\
    &=
    (\lambda-a)
    (\lambda-\beta_1)
    (\lambda-\beta_2)\\
    &~~
    -
    b_md_m(\lambda-\beta_2)
    -
    b_vd_v(\lambda-\beta_1).
\end{aligned}
\label{eq:app_characteristic_polynomial}
\end{equation}
The two loop gains
$b_md_m$ and $b_vd_v$
shift the poles of the joint system away from the uncoupled values
$a,\beta_1,\beta_2$.
The frozen system is asymptotically stable when every root of
Equation~\eqref{eq:app_characteristic_polynomial}
lies strictly inside the unit disk.

\subsection{Memory-to-Parameter Transfer Functions}
\label{app:frozen_transfer}

For a complex variable $z$ outside the spectrum of $\overline A$, define
\[
    \mathscr H_m(z)
    :=
    e_\theta^\top
    (zI-\overline A)^{-1}
    e_m,
\]
and
\[
    \mathscr H_v(z)
    :=
    e_\theta^\top
    (zI-\overline A)^{-1}
    e_v.
\]
Using the corresponding cofactors,
\begin{align*}
    \mathscr H_m(z)
    &=
    \frac{
        b_m(z-\beta_2)
    }{
        p(z)
    },\\
    \mathscr H_v(z)
    &=
    \frac{
        b_v(z-\beta_1)
    }{
        p(z)
    }.
\end{align*}
Both channels share the poles of the complete joint system. Their numerators
differ because the momentum input bypasses the $v$ state, whereas the
second-moment input bypasses the $m$ state.

\subsection{Modal Decomposition and Nonnormal Residues}
\label{app:modal_decomposition}

Suppose $\overline A$ is diagonalizable over $\mathbb C$:
\[
    \overline A
    =
    V\Lambda V^{-1},
\]
where
\[
    V
    =
    \begin{bmatrix}
        v_1&v_2&v_3
    \end{bmatrix},
    ~~
    V^{-1}
    =
    \begin{bmatrix}
        w_1^*\\
        w_2^*\\
        w_3^*
    \end{bmatrix},
\]
and
\[
    w_i^*v_j
    =
    \delta_{ij}.
\]
Then
\[
    \overline A^{h-1}
    =
    \sum_{i=1}^3
    \lambda_i^{h-1}
    v_iw_i^*.
\]
Therefore
\begin{equation}
\begin{aligned}
    r_k(h)
    &=
    e_\theta^\top
    \overline A^{h-1}
    e_k\\
    &=
    \sum_{i=1}^3
    \left(
        e_\theta^\top v_i
    \right)
    \left(
        w_i^*e_k
    \right)
    \lambda_i^{h-1}.
\end{aligned}
\label{eq:app_modal_response}
\end{equation}
For a real matrix, complex eigenvalues and residues occur in conjugate pairs,
so Equation~\eqref{eq:app_modal_response} remains real.

The modal expansion implies
\[
    |r_k(h)|
    \leq
    \sum_{i=1}^3
    \left|
        e_\theta^\top v_i
    \right|
    \left|
        w_i^*e_k
    \right|
    |\lambda_i|^{h-1}.
\]
A coarser matrix-norm bound is
\[
\begin{aligned}
    |r_k(h)|
    &\leq
    \|e_\theta^\top V\|_2
    \|\Lambda^{h-1}\|_2
    \|V^{-1}e_k\|_2\\
    &\leq
    \kappa_2(V)
    \rho(\overline A)^{h-1},
\end{aligned}
\]
where
\[
    \kappa_2(V)
    :=
    \|V\|_2\|V^{-1}\|_2.
\]

For a normal matrix, $V$ can be chosen unitary and
$\kappa_2(V)=1$.
For a nonnormal matrix, the eigenvector condition number and individual
input--output residues can be much larger. Nonnormality can therefore enlarge
the finite-horizon memory-to-parameter response even when all eigenmodes are
asymptotically decaying.

The finite-horizon peak
\[
    h_k^\star
    :=
    \min\arg\max_{h\geq1}|r_k(h)|
\]
depends jointly on the modal decay rates, oscillatory phases, and
input--output residues. Multiple decaying modes can interfere constructively at
intermediate horizons, placing the largest response after the initial
memory-to-parameter conversion. The sign of the corresponding probe-loss
response additionally depends on the scalar probe readout.

\subsection{Signs of the Two Feedback Loops}
\label{app:loop_signs}

In the scalar restriction,
\[
    b_m
    =
    \beta_1M,
    ~~
    d_m
    =
    (1-\beta_1)H^{\mathrm{tr}}.
\]
Since $M<0$,
\[
    \operatorname{sign}(b_md_m)
    =
    -\operatorname{sign}(H^{\mathrm{tr}}).
\]
Thus, positive local curvature gives a negative momentum-loop return gain,
which can contribute to oscillatory or sign-changing parameter responses.

For the second-moment loop,
\[
    b_v
    =
    \beta_2V,
    ~~
    d_v
    =
    2(1-\beta_2)gH^{\mathrm{tr}}.
\]
Since the sign of $V$ follows the sign of the current first-moment state,
\[
    \operatorname{sign}(b_vd_v)
    =
    \operatorname{sign}
    \left(
        m\,g\,H^{\mathrm{tr}}
    \right).
\]
The second-moment feedback can therefore reinforce or oppose the evolving
parameter response depending on the local operating point.

\subsection{Weight Decay and Bias Correction}
\label{app:frozen_wd_bias}

For scalar weight decay $\lambda_{\mathrm{wd}}$,
\[
    a
    =
    1-\eta\lambda_{\mathrm{wd}}
    +
    \Gamma H^{\mathrm{tr}}.
\]
Weight decay therefore modifies the direct parameter-retention term and,
through $a$, changes the short-horizon response coefficients and their
interaction with the two moment-memory channels.

Bias correction enters through
\[
    \rho_{1,s}
    =
    1-\beta_1^s,
    ~~
    \rho_{2,s}
    =
    1-\beta_2^s,
\]
and hence through $M_s$, $V_s$, and $\Gamma_s$.
In the full AdamW dynamics these quantities are time dependent. The frozen
model treats their values at the selected operating point as fixed over the
local analysis window. Bias correction therefore changes the numerical
coefficients of the frozen system without changing its parameter--moment
coupling structure.


\section{Finite-Horizon Approximation Error}
\label{app:fh_error}

This appendix proves
Theorem~\ref{thm:finite_horizon_error}
and
Corollary~\ref{cor:fixed_horizon_accuracy}.
It also gives a sufficient activation-margin condition for quadratic switching
error, a recursive error envelope, and the pathwise interpretation under
future training randomness.

\subsection{Exact Smooth--Switching Decomposition}
\label{app:smooth_switching_decomposition}

Let
\[
    x_s^{\mathrm s}
    :=
    x_s^{\mathrm c}
    +
    \delta x_s.
\]
The exact paired state difference after one future update is
\[
\begin{aligned}
    \delta x_{s+1}
    &=
    f_s
    \left(
        x_s^{\mathrm c}+\delta x_s
    \right)
    -
    f_s
    \left(
        x_s^{\mathrm c}
    \right).
\end{aligned}
\]
By construction,
\[
    f_s(x_s^{\mathrm c})
    =
    \widetilde f_s(x_s^{\mathrm c}).
\]
Adding and subtracting
$\widetilde f_s(x_s^{\mathrm c}+\delta x_s)$ gives
\[
\begin{aligned}
    \delta x_{s+1}
    ={}&
    \widetilde f_s
    \left(
        x_s^{\mathrm c}+\delta x_s
    \right)
    -
    \widetilde f_s
    \left(
        x_s^{\mathrm c}
    \right)\\
    &+
    f_s
    \left(
        x_s^{\mathrm c}+\delta x_s
    \right)
    -
    \widetilde f_s
    \left(
        x_s^{\mathrm c}+\delta x_s
    \right).
\end{aligned}
\]
Using
\begin{align*}
    r_s^{\mathrm{sm}}
    &:=
    \widetilde f_s
    \left(
        x_s^{\mathrm c}+\delta x_s
    \right)
    -
    \widetilde f_s
    \left(
        x_s^{\mathrm c}
    \right)
    -
    A_s\delta x_s,\\
    r_s^{\mathrm{sw}}
    &:=
    f_s
    \left(
        x_s^{\mathrm c}+\delta x_s
    \right)
    -
    \widetilde f_s
    \left(
        x_s^{\mathrm c}+\delta x_s
    \right),
\end{align*}
we obtain
\begin{equation}
    \delta x_{s+1}
    =
    A_s\delta x_s
    +
    r_s^{\mathrm{sm}}
    +
    r_s^{\mathrm{sw}}.
    \label{eq:app_exact_perturbation_recursion}
\end{equation}

If $D\widetilde f_s$ is $L_s$-Lipschitz along
\[
    \mathcal L_s
    =
    \left\{
        x_s^{\mathrm c}
        +
        \tau\delta x_s:
        0\leq\tau\leq1
    \right\},
\]
the integral remainder formula gives
\[
    r_s^{\mathrm{sm}}
    =
    \int_0^1
    \left[
        D\widetilde f_s
        \left(
            x_s^{\mathrm c}
            +
            \tau\delta x_s
        \right)
        -
        D\widetilde f_s(x_s^{\mathrm c})
    \right]
    \delta x_s
    \,\mathrm d\tau.
\]
Therefore
\begin{equation}
\begin{aligned}
    \|r_s^{\mathrm{sm}}\|
    &\leq
    \int_0^1
    L_s\tau
    \|\delta x_s\|^2
    \,\mathrm d\tau\\
    &=
    \frac{L_s}{2}
    \|\delta x_s\|^2.
\end{aligned}
\label{eq:app_smooth_remainder_bound}
\end{equation}

For a smooth network,
$r_s^{\mathrm{sw}}=0$.
The same holds in a piecewise-smooth network whenever the paired states remain
in the same activation region.

\subsection{Proof of the Multistep State-Error Identity}
\label{app:multistep_error_proof}

The tangent approximation is initialized by
\[
    \widehat{\delta x}_{t+1}
    :=
    B_t\xi_t
\]
and propagated according to
\[
    \widehat{\delta x}_{s+1}
    :=
    A_s\widehat{\delta x}_s.
\]
Define
\[
    e_s
    :=
    \delta x_s-\widehat{\delta x}_s.
\]
At the first post-shock state,
\begin{equation}
\begin{aligned}
    e_{t+1}
    &=
    \delta x_{t+1}
    -
    \widehat{\delta x}_{t+1}\\
    &=
    B_t\xi_t
    +
    r_{B,t}(\xi_t)
    -
    B_t\xi_t\\
    &=
    r_{B,t}(\xi_t).
\end{aligned}
\label{eq:app_initial_error}
\end{equation}

For a future step, subtracting the tangent recursion from
Equation~\eqref{eq:app_exact_perturbation_recursion} gives
\[
\begin{aligned}
    e_{s+1}
    &=
    A_s\delta x_s
    +
    r_s^{\mathrm{sm}}
    +
    r_s^{\mathrm{sw}}
    -
    A_s\widehat{\delta x}_s\\
    &=
    A_se_s
    +
    r_s^{\mathrm{sm}}
    +
    r_s^{\mathrm{sw}}.
\end{aligned}
\]
Repeated substitution yields
\begin{equation}
\begin{aligned}
    e_{t+h}
    ={}&
    \Phi_{t+h,t+1}
    r_{B,t}(\xi_t)\\
    &+
    \sum_{j=t+1}^{t+h-1}
    \Phi_{t+h,j+1}
    \left(
        r_j^{\mathrm{sm}}
        +
        r_j^{\mathrm{sw}}
    \right).
\end{aligned}
\label{eq:app_exact_multistep_error}
\end{equation}
For $h=1$, the sum is empty and
Equation~\eqref{eq:app_exact_multistep_error}
reduces to
Equation~\eqref{eq:app_initial_error}.

Taking norms gives
\[
\begin{aligned}
    \|e_{t+h}\|
    \leq{}&
    \|\Phi_{t+h,t+1}\|
    \|r_{B,t}(\xi_t)\|\\
    &+
    \sum_{j=t+1}^{t+h-1}
    \|\Phi_{t+h,j+1}\|
    \left(
        \|r_j^{\mathrm{sm}}\|
        +
        \|r_j^{\mathrm{sw}}\|
    \right).
\end{aligned}
\]
Using
Equation~\eqref{eq:app_smooth_remainder_bound}
proves the state-error bound in
Theorem~\ref{thm:finite_horizon_error}.

\subsection{Probe-Output Error}
\label{app:output_error}

Let
\[
    \Pi_\theta
    :=
    \begin{bmatrix}
        I&0&0
    \end{bmatrix}
\]
denote projection onto the parameter block. Then
\[
    \delta\theta_{t+h}
    =
    \Pi_\theta\delta x_{t+h}.
\]
The exact paired probe response is
\[
\begin{aligned}
    \delta F_{t+h}
    =
    F_{t+h}
    \left(
        \theta_{t+h}^{\mathrm c}
        +
        \delta\theta_{t+h}
    \right)
    -
    F_{t+h}
    \left(
        \theta_{t+h}^{\mathrm c}
    \right).
\end{aligned}
\]
If
$\nabla F_{t+h}$
is $L_{F,t+h}$-Lipschitz along the connecting segment, then
\[
    \delta F_{t+h}
    =
    c_{t+h}^{\top}
    \delta x_{t+h}
    +
    r_{F,t+h},
\]
where
\begin{equation}
    |r_{F,t+h}|
    \leq
    \frac{L_{F,t+h}}{2}
    \|\delta\theta_{t+h}\|^2.
    \label{eq:app_output_remainder}
\end{equation}

The tangent output prediction is
\[
\begin{aligned}
    \widehat{\delta F}_{t+h}
    &=
    c_{t+h}^{\top}
    \widehat{\delta x}_{t+h}\\
    &=
    c_{t+h}^{\top}
    \Phi_{t+h,t+1}
    B_t\xi_t.
\end{aligned}
\]
Therefore
\[
\begin{aligned}
    \delta F_{t+h}
    -
    \widehat{\delta F}_{t+h}
    &=
    c_{t+h}^{\top}
    \left(
        \delta x_{t+h}
        -
        \widehat{\delta x}_{t+h}
    \right)
    +
    r_{F,t+h}\\
    &=
    c_{t+h}^{\top}e_{t+h}
    +
    r_{F,t+h}.
\end{aligned}
\]
Taking absolute values yields
\[
\begin{aligned}
    \left|
        \delta F_{t+h}
        -
        c_{t+h}^{\top}
        \Phi_{t+h,t+1}
        B_t\xi_t
    \right|
    \leq{}&
    \|c_{t+h}\|
    \|e_{t+h}\|\\
    &+
    \frac{L_{F,t+h}}{2}
    \|\delta\theta_{t+h}\|^2,
\end{aligned}
\]
which proves the output-error bound in
Theorem~\ref{thm:finite_horizon_error}.

\subsection{A Sufficient Activation-Margin Condition}
\label{app:activation_switching}

We now give a sufficient condition under which the aggregate switching defect
satisfies the quadratic bound required by
Corollary~\ref{cor:fixed_horizon_accuracy}.

Let
$\mathcal S_s$
denote the set of activation gates whose states differ between the paired
trajectories during transition $s$.
For multiple simultaneous switches, the total switching defect can be
decomposed by a telescoping construction.

Choose an arbitrary ordering
\[
    \mathcal S_s
    =
    \{j_1,\ldots,j_q\}.
\]
For $r=0,\ldots,q$, let $f_s^{(r)}$ denote the local transition map that uses
the shock activation state for gates $j_1,\ldots,j_r$ and the control
activation state for the remaining switched gates. At the perturbed state,
\[
    f_s^{(0)}
    =
    \widetilde f_s,
    ~~
    f_s^{(q)}
    =
    f_s.
\]
Define
\[
\begin{aligned}
    r_{s,j_r}^{\mathrm{sw}}
    :={}&
    f_s^{(r)}
    \left(
        x_s^{\mathrm c}+\delta x_s
    \right)\\
    &-
    f_s^{(r-1)}
    \left(
        x_s^{\mathrm c}+\delta x_s
    \right).
\end{aligned}
\]
The total switching defect telescopes:
\[
    r_s^{\mathrm{sw}}
    =
    \sum_{r=1}^{q}
    r_{s,j_r}^{\mathrm{sw}}.
\]

Assume that each local switch satisfies
\begin{equation}
    \|r_{s,j}^{\mathrm{sw}}\|
    \leq
    K_{s,j}\|\delta x_s\|.
    \label{eq:app_per_switch_bound}
\end{equation}
Let
$a_{s,j}^{\mathrm c}$
denote the control pre-activation of gate $j$, and let
$\delta a_{s,j}$ denote its change between the paired states. A gate can switch
only if
\[
    |a_{s,j}^{\mathrm c}|
    \leq
    |\delta a_{s,j}|.
\]
Assume
\[
    |\delta a_{s,j}|
    \leq
    L_{a,s,j}\|\delta x_s\|,
    ~~
    L_{a,s,j}
    \leq
    \overline L_{a,s}.
\]
Every switched unit therefore satisfies
\[
    |a_{s,j}^{\mathrm c}|
    \leq
    \overline L_{a,s}
    \|\delta x_s\|,
\]
and hence
\begin{equation}
    \mathcal S_s
    \subseteq
    \left\{
        j:
        |a_{s,j}^{\mathrm c}|
        \leq
        \overline L_{a,s}
        \|\delta x_s\|
    \right\}.
    \label{eq:app_switch_set}
\end{equation}

Suppose further that, for all sufficiently small $u>0$, the weighted mass of
units near the activation boundary satisfies
\begin{equation}
    \sum_{
        j:
        |a_{s,j}^{\mathrm c}|
        \leq u
    }
    K_{s,j}
    \leq
    \rho_su.
    \label{eq:app_weighted_margin_mass}
\end{equation}
Using
Equations~\eqref{eq:app_per_switch_bound}
and
\eqref{eq:app_switch_set},
\[
\begin{aligned}
    \|r_s^{\mathrm{sw}}\|
    &\leq
    \sum_{j\in\mathcal S_s}
    \|r_{s,j}^{\mathrm{sw}}\|\\
    &\leq
    \|\delta x_s\|
    \sum_{j\in\mathcal S_s}
    K_{s,j}\\
    &\leq
    \|\delta x_s\|
    \sum_{
        j:
        |a_{s,j}^{\mathrm c}|
        \leq
        \overline L_{a,s}
        \|\delta x_s\|
    }
    K_{s,j}.
\end{aligned}
\]
Applying
Equation~\eqref{eq:app_weighted_margin_mass}
with
\[
    u
    =
    \overline L_{a,s}
    \|\delta x_s\|
\]
gives
\begin{equation}
    \|r_s^{\mathrm{sw}}\|
    \leq
    \rho_s
    \overline L_{a,s}
    \|\delta x_s\|^2.
    \label{eq:app_switching_quadratic_bound}
\end{equation}
Thus the activation-margin condition provides the quadratic switching bound
used by the fixed-horizon accuracy result.

\subsection{Proof of Fixed-Horizon First-Order Accuracy}
\label{app:fixed_horizon_accuracy}

Consider the scaled perturbation
$\alpha\xi_t$
and a fixed horizon $H$.

Local Lipschitz continuity of $D\psi_t$ gives, from
Equation~\eqref{eq:app_writein_quadratic_bound},
\[
    \|r_{B,t}(\alpha\xi_t)\|
    \leq
    C_{B,t}
    \alpha^2
    \|\xi_t\|^2
\]
for sufficiently small $\alpha$.

Local Lipschitz continuity of
$D\widetilde f_s$
gives
\[
    \|r_s^{\mathrm{sm}}\|
    \leq
    \frac{L_s}{2}
    \|\delta x_s\|^2.
\]
Under the switching assumption in
Corollary~\ref{cor:fixed_horizon_accuracy},
\[
    \|r_s^{\mathrm{sw}}\|
    \leq
    C_s^{\mathrm{sw}}
    \|\delta x_s\|^2.
\]
Hence
\begin{equation}
    \|r_s^{\mathrm{sm}}\|
    +
    \|r_s^{\mathrm{sw}}\|
    \leq
    C_s
    \|\delta x_s\|^2,
    ~~
    C_s
    :=
    \frac{L_s}{2}
    +
    C_s^{\mathrm{sw}}.
    \label{eq:app_total_quadratic_remainder}
\end{equation}

At the first post-shock state,
\[
\begin{aligned}
    \|\delta x_{t+1}(\alpha)\|
    &\leq
    |\alpha|
    \|B_t\xi_t\|
    +
    C_{B,t}
    \alpha^2
    \|\xi_t\|^2\\
    &=
    O(|\alpha|).
\end{aligned}
\]
Suppose inductively that
\[
    \|\delta x_s(\alpha)\|
    =
    O(|\alpha|)
\]
for some $t+1\leq s<t+H$.
The local derivatives are bounded on the relevant neighborhood, so
$\|A_s\|$ is bounded. Using
Equation~\eqref{eq:app_exact_perturbation_recursion},
\[
\begin{aligned}
    \|\delta x_{s+1}(\alpha)\|
    &\leq
    \|A_s\|
    \|\delta x_s(\alpha)\|
    +
    C_s
    \|\delta x_s(\alpha)\|^2\\
    &=
    O(|\alpha|)
    +
    O(\alpha^2)\\
    &=
    O(|\alpha|).
\end{aligned}
\]
By induction,
\[
    \|\delta x_{t+h}(\alpha)\|
    =
    O(|\alpha|)
\]
for every fixed
$1\leq h\leq H$.

The exact multistep error identity now gives
\[
\begin{aligned}
    \|e_{t+h}(\alpha)\|
    \leq{}&
    \|\Phi_{t+h,t+1}\|
    O(\alpha^2)\\
    &+
    \sum_{j=t+1}^{t+h-1}
    \|\Phi_{t+h,j+1}\|
    O(\alpha^2).
\end{aligned}
\]
Since $H$ is fixed and the finite-horizon propagators are uniformly bounded,
\[
    \|e_{t+h}(\alpha)\|
    =
    O(\alpha^2),
    ~~
    1\leq h\leq H.
\]

The probe objectives have locally Lipschitz gradients, and
$\|\delta\theta_{t+h}(\alpha)\|=O(|\alpha|)$, so
Equation~\eqref{eq:app_output_remainder} gives
\[
    |r_{F,t+h}(\alpha)|
    =
    O(\alpha^2).
\]
Therefore
\[
\begin{aligned}
    \delta F_{t+h}(\alpha)
    &=
    c_{t+h}^{\top}
    \widehat{\delta x}_{t+h}(\alpha)
    +
    O(\alpha^2)\\
    &=
    \alpha
    c_{t+h}^{\top}
    \Phi_{t+h,t+1}
    B_t\xi_t
    +
    O(\alpha^2),
\end{aligned}
\]
for every
$1\leq h\leq H$.
The constants may depend on the fixed control trajectory and horizon $H$ but
not on $\alpha$. This proves
Corollary~\ref{cor:fixed_horizon_accuracy}.

When the sufficient activation-margin condition of
Appendix~\ref{app:activation_switching}
holds,
Equation~\eqref{eq:app_total_quadratic_remainder}
can be instantiated with
\[
    C_s^{\mathrm{sw}}
    =
    \rho_s\overline L_{a,s}.
\]

\subsection{Recursive Error Envelope}
\label{app:recursive_envelope}

The structural error bound contains the true deviations
$\delta x_s$.
A recursive envelope can instead be expressed in terms of the tangent
trajectory and previously accumulated error bounds.

Suppose
\[
    \|e_s\|
    \leq
    \varepsilon_s.
\]
Since
\[
    \delta x_s
    =
    \widehat{\delta x}_s+e_s,
\]
we have
\[
    \|\delta x_s\|
    \leq
    \|\widehat{\delta x}_s\|
    +
    \varepsilon_s.
\]
Using the error recursion,
\[
\begin{aligned}
    \|e_{s+1}\|
    &\leq
    \|A_s\|
    \|e_s\|
    +
    \|r_s^{\mathrm{sm}}\|
    +
    \|r_s^{\mathrm{sw}}\|\\
    &\leq
    \|A_s\|\varepsilon_s
    +
    \frac{L_s}{2}
    \left(
        \|\widehat{\delta x}_s\|
        +
        \varepsilon_s
    \right)^2
    +
    \|r_s^{\mathrm{sw}}\|.
\end{aligned}
\]
Thus the recursion
\begin{equation}
\begin{aligned}
    \varepsilon_{s+1}
    :={}&
    \|A_s\|\varepsilon_s\\
    &+
    \frac{L_s}{2}
    \left(
        \|\widehat{\delta x}_s\|
        +
        \varepsilon_s
    \right)^2
    +
    \|r_s^{\mathrm{sw}}\|
\end{aligned}
\label{eq:app_epsilon_recursion}
\end{equation}
preserves
\[
    \|e_{s+1}\|
    \leq
    \varepsilon_{s+1}.
\]
Initializing with
\[
    \varepsilon_{t+1}
    :=
    \|r_{B,t}(\xi_t)\|
\]
therefore gives an envelope for all subsequent states.

If a quadratic switching bound
\[
    \|r_s^{\mathrm{sw}}\|
    \leq
    C_s^{\mathrm{sw}}
    \|\delta x_s\|^2
\]
is available, then
\[
    \|r_s^{\mathrm{sw}}\|
    \leq
    C_s^{\mathrm{sw}}
    \left(
        \|\widehat{\delta x}_s\|
        +
        \varepsilon_s
    \right)^2.
\]
The envelope becomes
\begin{equation}
\begin{aligned}
    \varepsilon_{s+1}
    :={}&
    \|A_s\|\varepsilon_s\\
    &+
    \left(
        \frac{L_s}{2}
        +
        C_s^{\mathrm{sw}}
    \right)
    \left(
        \|\widehat{\delta x}_s\|
        +
        \varepsilon_s
    \right)^2.
\end{aligned}
\label{eq:app_quadratic_epsilon_recursion}
\end{equation}
Under the activation-margin sufficient condition,
\[
    C_s^{\mathrm{sw}}
    =
    \rho_s\overline L_{a,s}.
\]

The same transition dynamics therefore govern both the desired first-order
response and the accumulation of approximation error: large finite-horizon
gain amplifies the propagated perturbation as well as nonlinear defects
introduced along the trajectory.

\subsection{Pathwise Conditioning and Distributional Extension}
\label{app:pathwise_conditioning}

Let $\omega$ denote a realized future random sequence, including future
minibatches, dropout masks, data augmentation, and other stochastic training
operations. Conditional on $\omega$ and the pre-shock history, the future
operators are deterministic:
\[
    A_s(\omega),
    ~~
    \Phi_{t+h,t+1}(\omega),
    ~~
    c_{t+h}(\omega).
\]
When the probe sequence is itself defined from the future minibatches, such as
\[
    F_s(\theta)
    =
    \ell(\theta;\mathcal B_s),
\]
the probe functions are also fixed after conditioning on $\omega$.

The corresponding pathwise input--output operator is
\[
    \mathcal G_{t,h}(\omega)
    =
    c_{t+h}(\omega)^\top
    \Phi_{t+h,t+1}(\omega)
    B_t.
\]
All directional-response identities and error decompositions above apply
separately to each realized common-future sequence. Averaging a pathwise
quantity over independently sampled futures yields its corresponding
distributional version. For example,
\[
    \mathbb E_\omega
    \left[
        \delta F_{t+h}(\omega)
    \right]
\]
describes the mean paired response at horizon $h$, while
\[
    \mathbb E_\omega
    \left[
        \mathcal G_{t,h}(\omega)\xi_t
    \right]
\]
gives the corresponding first-order mean response whenever the expectation and
local expansion may be interchanged.

The common-future construction retains the same future realization for the
control and perturbed trajectories within each pair. It therefore isolates the
pathwise propagation associated with the perturbation at step $t$, while
repetition across different future realizations characterizes variability of
that response.

\section{Detailed Experimental Protocols and Results}
\label{app:experimental_details}

This appendix gives the complete protocols and the additional numerical results
supporting Section~\ref{sec:experiments}.  The experiments are organized by
scientific question rather than by model class.  Experiment~1 validates the
finite-horizon mechanism across controlled quadratic systems, nonconvex neural
networks, and pretrained language models.  Experiment~2 then asks whether the
resulting delayed response retains candidate-specific structure when the future
training sequence is unknown and independently resampled.

For the controlled and neural-network mechanism-validation stages, horizon
$h=1$ denotes the first post-shock state and
\begin{align}
    d_{i,h}(\alpha)
    &:=F(\theta^{\mathrm{s}}_{i,t+h}(\alpha))
    -F(\theta^{\mathrm{c}}_{t+h}),\\
    \widehat d_{i,h}
    &:=c_{t+h}^{\top}\Phi_{t+h,t+1}B_t\xi_i,\\
    \widehat d_{i,h}(\alpha)
    &:=\alpha\widehat d_{i,h}.
\end{align}
We use
\begin{align}
    M_i(\alpha)&=\max_h|d_{i,h}(\alpha)|,
    &\operatorname{ARE}_i(\alpha)&=\sum_h|d_{i,h}(\alpha)|,\\
    h_i^\star&=\min\arg\max_h|d_{i,h}|,
    &s_i^\star&=\operatorname{sign}(d_{i,h_i^\star}).
\end{align}
The primary trajectory error is
\begin{equation}
\label{eq:app_nrmse}
    \operatorname{NRMSE}_{i,\alpha}
    =
    \left(
    \frac{\sum_h(d_{i,h}(\alpha)-\alpha\widehat d_{i,h})^2}
         {\sum_h d_{i,h}(\alpha)^2}
    \right)^{1/2}.
\end{equation}
For the analytic controlled and neural-network tangents, the horizon-wise
symmetric relative error is
\begin{equation}
\label{eq:app_relative_error}
    e_{i,h}(\alpha)
    =
    \frac{|d_{i,h}(\alpha)-\alpha\widehat d_{i,h}|}
    {|d_{i,h}(\alpha)|+|\alpha\widehat d_{i,h}|+10^{-30}}.
\end{equation}
The empirical validity radius is defined on the tested scale grid by
\begin{equation}
\label{eq:app_validity_radius}
    \alpha_i^{\mathrm{valid}}
    =
    \max\left\{\alpha:
    \operatorname{median}_{1\leq h\leq H}e_{i,h}(\alpha)\leq0.2
    \right\},
\end{equation}
with value zero when the set is empty.  Candidate radii are aggregated within
each system before system-level summaries are formed.  The language-model
stage reports the corresponding valid-candidate fraction at every tested
scale.

Candidates, scales, horizons, and future branches are always treated as nested
observations.  The controlled and neural-network studies first aggregate
candidate-level quantities within independently generated training systems.
The Pythia study is a single-seed scaling analysis: its nine model--dataset
systems are fixed conditions rather than independent random replications, and
its statistics are descriptive.  Experiment~2 preserves these system units
while introducing 32 independently resampled future continuations per fixed
state--shock pair.

\subsection{Common Baselines and Tangent Ablations}
\label{app:baselines}

The controlled and neural-network stages of Experiment~1 explicitly construct the initial AdamW tangent
\begin{equation}
    \delta x_{i,1}
    :=B_t\xi_i
    =\begin{bmatrix}
        \delta\theta_{i,t+1}\\
        \delta m_{i,t}\\
        \delta v_{i,t}
      \end{bmatrix}
\end{equation}
and recursively propagate
$\delta x_{i,h+1}=A_{t+h}\delta x_{i,h}$.
Let $P_\theta\delta x=\delta\theta$ and use the local shorthand
$\delta\theta_{i,h}:=P_\theta\delta x_{i,h}$. Define the parameter-space
probe gradient
\begin{equation}
    \bar c_h
    :=\nabla F(\theta^{\mathrm c}_{t+h}),
    \qquad
    c_{t+h}=P_\theta^\top\bar c_h.
\end{equation}
Every trajectory score below is the maximum absolute predicted response over
$1\leq h\leq H$.

\paragraph{Full ISO.}
The complete analytic/JVP trajectory and its score are
\begin{equation}
\label{eq:app_full_iso}
    r_{i,h}^{\mathrm{ISO}}
    =\bar c_h^\top P_\theta\Phi_{t+h,t+1}B_t\xi_i,
    \qquad
    S_i^{\mathrm{ISO}}=\max_h|r_{i,h}^{\mathrm{ISO}}|.
\end{equation}

\paragraph{Temporal ablations.}
The implementations used in both mechanism-validation stages are
\begin{align}
\label{eq:app_temporal_ablations}
    r_{i,h}^{\mathrm{no\text{-}prop}}
    &=\bar c_h^\top\delta\theta_{i,1},\\
    r_{i,h}^{\mathrm{init\text{-}\theta}}
    &=\bar c_h^\top P_\theta\Phi_{t+h,t+1}
      P_\theta^\top\delta\theta_{i,1},\\
    r_{i,h}^{\mathrm{clamp\text{-}\theta}}
    &=\bar c_h^\top
      \left(\prod_{s=t+1}^{t+h-1}[A_s]_{\theta\theta}\right)
      \delta\theta_{i,1},\\
    r_{i,h}^{\mathrm{frozen\text{-}dyn}}
    &=\bar c_h^\top P_\theta A_{t+1}^{h-1}\delta x_{i,1},\\
    r_{i,h}^{\mathrm{frozen\text{-}readout}}
    &=\bar c_1^\top\delta\theta_{i,h}.
\end{align}
Here, \textsc{No Propagation} reuses the initial parameter write at every
horizon.  \textsc{Initial Parameter Only} removes the two moment components of
$\delta x_{i,1}$ once and then applies the complete time-varying dynamics.  \textsc{Clamped
Parameter State} sets the moment tangent blocks to zero before and after every
transition, retaining only $[A_s]_{\theta\theta}$.
\textsc{Frozen Dynamics} repeatedly applies the first future Jacobian, and
\textsc{Frozen Readout} applies $\bar c_1$ to the correctly propagated parameter
tangent at every horizon.

\paragraph{Scalar baselines.}
The exact immediate-response oracle and the static scores are
\begin{align}
\label{eq:app_scalar_baselines}
    S_i^{\mathrm{1step}}&=|d_{i,1}(1)|,
    &S_i^{\mathrm{grad}}&=\|\xi_i\|_2,\\
    S_i^{\mathrm{write}}&=\|\delta\theta_{i,1}\|_2,
    &S_i^{\mathrm{curv}}&=|\xi_i^\top H_{\mathrm{ref}}\xi_i|,\\
    S_i^{\mathrm{norm}}&=\max_h\|\bar c_h\|_2\|\delta\theta_{i,h}\|_2,\\
    H_{\mathrm{ref}}&=\frac{1}{R}\sum_{r=1}^R
      \nabla^2\ell(\theta_t;\mathcal B_r^{\mathrm{ref}}).
\end{align}
The curvature score therefore uses the shock direction in the average
reference-minibatch Hessian.  The norm-product baseline removes directional
alignment while retaining the horizon-wise state and readout norms.  For the
readout-order null, one shared permutation $\pi$ is applied to all candidates
in a system and
\begin{equation}
    S_{i,\pi}^{\mathrm{shuffle}}
    =\max_h|\bar c_{\pi(h)}^\top\delta\theta_{i,h}|.
\end{equation}
We use 100 independently drawn permutations per system.

\subsection{Experiment 1: Finite-Horizon Mechanism Validation}
\label{app:exp1_mechanism}

\subsubsection{Controlled Quadratic Systems}
\label{app:exp1_controlled_details}
\label{app:exp1_details}

\paragraph{Systems and protocol.}
Each minibatch loss is
\begin{equation}
\label{eq:app_exp1_quadratic}
    \ell_s(\theta)
    =\frac12\theta^\top D_s\theta
    +\frac{1}{2r}\|U_s^\top\theta\|_2^2+q_s^\top\theta,
\end{equation}
where $D_s$ is positive diagonal, $U_s\in\mathbb R^{d\times r}$,
$d=512$, and $r=16$.  For condition multiplier
$\kappa\in\{1,4,16\}$, independent innovations are sampled as
\begin{align}
    \widetilde D_{s,j}
    &\sim\operatorname{Unif}(0.05,0.25\kappa),\\
    \widetilde U_{s,jk}
    &\sim\mathcal N(0,1/d),\\
    \widetilde q_{s,j}
    &\sim\mathcal N(0,0.05^2).
\end{align}
Writing $\rho=0.85$, the temporally correlated sequence is generated by
\begin{align}
    D_s&=\rho D_{s-1}+(1-\rho)\widetilde D_s,\\
    U_s&=\rho U_{s-1}+\sqrt{1-\rho^2}\,\widetilde U_s,\\
    q_s&=\rho q_{s-1}+\sqrt{1-\rho^2}\,\widetilde q_s,
\end{align}
with the first batch equal to its innovation.  The condition multiplier cycles
through $\{1,4,16\}$ with the system index, and
$\theta_0\sim\mathcal N(0,0.1^2I)$.  AdamW uses
\begin{equation}
\begin{gathered}
    \eta=2\times10^{-3},\quad \beta_1=0.9,\quad \beta_2=0.999,\\
    \epsilon=10^{-8},\quad\lambda=0.01.
\end{gathered}
\end{equation}
All calculations use double precision.  Each independently initialized system
is advanced for 40 burn-in steps.  Four reference minibatches define the
control gradient, 16 candidate minibatches define shocks, and a shared future
sequence is followed for $H=32$.  Four seeds and eight systems per seed give
32 independent base systems.  Standard and anisotropic probes are paired
views of each base system and are not counted as separate replications.

\paragraph{Probe constructions.}
The main controlled experiment uses two paired probes.  The
\emph{standard probe} is sampled independently from the same quadratic family,
with linear standard deviation $0.02$.  The \emph{anisotropic probe} starts
from that draw, sets its linear term to zero, and replaces its diagonal by
\begin{align}
    D_F^{\mathrm{aniso}}&=D_F\operatorname{Diag}(w),\\
    \{w_j\}_{j=1}^d
    &=\operatorname{perm}\!\left(
      \left\{32^{(j-1)/(d-1)}\right\}_{j=1}^d
    \right),
\end{align}
while retaining the sampled low-rank factor.  This is the
\texttt{readout\_active} regime in the code.

The exact-one-step-matched supplement uses the standard probe and a separate
\emph{rotating-readout probe}.  Let $\theta^{\mathrm c}_{h_m}$ and
$\theta^{\mathrm c}_{h_m+1}$ be the two central states of the $H$-step common
control trajectory and set
\begin{align}
    \mu&=\frac12\left(
      \theta^{\mathrm c}_{h_m}+\theta^{\mathrm c}_{h_m+1}
    \right),\\
    W&=\operatorname{Diag}\!\left(
      \operatorname{perm}\!\left(
      \left\{32^{(j-1)/(d-1)}\right\}_{j=1}^d\right)
    \right).
\end{align}
The fixed probe is
\begin{equation}
\label{eq:app_rotating_probe}
    F_{\mathrm{rot}}(\theta)
    =\frac12(\theta-\mu)^\top W(\theta-\mu),
    \qquad
    \bar c_h=W(\theta_h^{\mathrm c}-\mu).
\end{equation}
It depends only on the common control trajectory.  Centering it between the
two middle states makes the readout direction change as the control trajectory
passes the probe center.  The standard, anisotropic, and rotating-readout
probes are therefore three distinct constructions.

\paragraph{Local fidelity and finite-scale error.}
We use
$\alpha\in\{1/32,1/16,1/8,1/4,1/2,1\}$.  Table
\ref{tab:app_exp1_fidelity} reports system-level medians.  At the smallest
scale, NRMSE is $0.0117$ under the standard probe and $0.0173$ under the
anisotropic probe.  The median local trajectory NRMSE pooled across the
prespecified local range is $0.0450$, and median horizon-wise sign agreement
is one.  The median empirical validity radius is $0.5$.  Fitting
\begin{equation}
    \log|d_{i,h}(\alpha)-\alpha\widehat d_{i,h}|
    =a_{i,h}+p_{i,h}\log\alpha
\end{equation}
over the local scales yields an approximately quadratic remainder, with
system-level $R^2$ values essentially one.  The independently evaluated
recursive state-error identity has maximum residual
$3.11\times10^{-17}$, providing an implementation audit of the recursive
decomposition.

\begin{table}[ht]
\centering
\caption{Controlled-stage signed-trajectory fidelity in Experiment~1.  Entries are medians over
32 independent systems after candidate-level aggregation.}
\label{tab:app_exp1_fidelity}
\small
\begin{tabular}{ccccc}
\toprule
Probe & $\alpha$ & NRMSE & Rel. $M$ err. & Sign acc.\\
\midrule
Standard & $1/32$ & 0.0117 & 0.0052 & 1.000\\
         & $1/8$  & 0.0483 & 0.0216 & 1.000\\
         & $1/2$  & 0.2256 & 0.0945 & 1.000\\
         & $1$    & 0.5538 & 0.2006 & 0.938\\
\midrule
Anisotropic & $1/32$ & 0.0173 & 0.0071 & 1.000\\
            & $1/8$  & 0.0719 & 0.0289 & 1.000\\
            & $1/2$  & 0.3445 & 0.1300 & 0.938\\
            & $1$    & 0.9512 & 0.2681 & 0.875\\
\bottomrule
\end{tabular}
\end{table}

\paragraph{Natural candidate ranking.}
For unmodified candidate minibatches, Full ISO reaches median Spearman
correlation $0.835$ for the standard probe and $0.762$ for the anisotropic
probe.  The corresponding exact one-step correlations are $0.774$ and
$0.734$.  Their difference is modest because natural immediate and future responses
can be strongly correlated.  Static
gradient, parameter-write, and curvature summaries attain correlations in
the approximate range $0.33$--$0.48$.

\paragraph{Exact-one-step matching.}
Let $a$ be the one-step tangent functional satisfying
$a^\top\xi=\bar c_1^\top P_\theta B_t\xi$.  In the implementation, the diagonal
write-in map is recovered by applying $B_t$ to the all-ones direction, so
$a=\bar c_1\odot P_\theta B_t\mathbf 1$.  For natural residual
$r_i=g_i-\bar g$, define
\begin{align}
    z_i&=r_i-a\frac{a^\top r_i}{\|a\|_2^2},\\
    b_i&=s_i\mu_1\frac{a}{\|a\|_2^2},
    \qquad s_i=(-1)^i,\\
    \widetilde z_i&=
    \begin{cases}
      4\|b_i\|_2z_i/\|z_i\|_2,&\|z_i\|_2>0,\\
      0,&\text{otherwise},
    \end{cases}\\
    \widetilde\xi_i&=b_i+\widetilde z_i.
\end{align}
The common target is
\begin{equation}
    \mu_1=0.25\operatorname{median}_j|a^\top r_j|.
\end{equation}
Thus $a^\top\widetilde\xi_i=s_i\mu_1$, while the candidate-specific
component lies in the nullspace of the immediate output functional and has
four times the norm of the common component.

We then calibrate one scalar $\gamma_i\geq0$ per candidate using the exact
nonlinear AdamW update:
Let $\Theta^+(x,g)$ denote the parameter component after one exact AdamW
update from state $x$ with gradient $g$.  The calibration equation is
\begin{equation}
\label{eq:app_exact_matching}
    F\!\left(\Theta^+(x_t,\bar g+\gamma_i\widetilde\xi_i)\right)
    -F\!\left(\Theta^+(x_t,\bar g)\right)
    =s_i\mu_1.
\end{equation}
Starting with $[0,1]$, the upper endpoint is doubled until the target is
bracketed, up to $\gamma=64$; 64 bisection iterations are then applied.  A
match is accepted when the signed residual is at most
$5\times10^{-8}\mu_1$.  All candidates across the 64 system--probe instances
(32 systems evaluated under two probe constructions) are successfully
calibrated.  This procedure uses the control state, the
one-step functional, and natural residual directions, but no response at
$h>1$.

Table~\ref{tab:app_exp1_matching} shows that the matched immediate response is
numerically constant whereas future magnitudes remain heterogeneous.  The
median ratio $M_i/|d_{i,1}|$ is $6.21$ under the standard probe and $4.54$
under the rotating-readout probe; corresponding parameter transient gains are
approximately $7.7$.

\begin{table}[ht]
\centering
\caption{Controlled-stage exact-one-step-matched stress test in Experiment~1.  CV is computed
over 16 candidates within a system; entries are medians across 32 independent
systems, with IQRs shown for the matching diagnostics.}
\label{tab:app_exp1_matching}
\small
\setlength{\tabcolsep}{3pt}
\begin{tabular}{lcc}
\toprule
Quantity & Standard & Rotating-readout\\
\midrule
Calibration success & $1.000\ [1.000,1.000]$ & $1.000\ [1.000,1.000]$\\
CV of $|d_1|$ & $1.10\!\times\!10^{-12}$ & $6.70\!\times\!10^{-12}$\\
CV of future $M$ & $0.092\ [0.064,0.114]$ & $0.459\ [0.292,0.530]$\\
$M/|d_1|$ & $6.21\ [5.69,7.86]$ & $4.54\ [3.38,5.08]$\\
Parameter gain & $7.71\ [7.10,8.90]$ & $7.69\ [7.06,8.83]$\\
\bottomrule
\end{tabular}
\end{table}

\begin{table}[ht]
\centering
\caption{Median Spearman correlation under exact-one-step matching.  The
one-step score is tied by construction.}
\label{tab:app_exp1_matched_ablation}
\small
\setlength{\tabcolsep}{3pt}
\begin{tabular}{lcc}
\toprule
Method & Standard & Rotating-readout\\
\midrule
Full ISO & \textbf{0.993} & \textbf{1.000}\\
No propagation & 0.354 & 0.806\\
Frozen dynamics & 0.788 & 0.969\\
Initial parameter only & 0.137 & 0.788\\
Clamped parameter state & 0.254 & 0.793\\
Frozen readout & 0.371 & 0.831\\
Gradient norm & -0.062 & 0.776\\
Parameter-write norm & -0.025 & 0.647\\
Curvature & 0.044 & 0.738\\
Norm product & 0.041 & 0.790\\
\bottomrule
\end{tabular}
\end{table}

\paragraph{Readout controls.}
The standard readout remains close to its initial direction, with median
minimum cosine $0.914$.  The rotating-readout probe produces
substantial horizon-dependent rotation while keeping a fixed scalar probe
objective.  We additionally apply 100 common permutations to the readout
sequence while preserving the propagated states.  Correct temporal alignment
places Full ISO at the maximum observed percentile of the shuffled-readout
null in both regimes.  This control distinguishes directional state--readout
alignment from a product of state and readout norms.

\paragraph{State-channel dependence and persistence.}
Let the exact post-shock joint-state deviation be
\begin{equation}
    \Delta x_{t+1}
    =
    (\Delta\theta_{t+1},\Delta m_t,\Delta v_t).
\end{equation}
For $(a,b,c)\in\{0,1\}^3$, we form
\begin{equation}
    x_{t+1}^{abc}=x_{t+1}^{\mathrm c}
      +(a\Delta\theta_{t+1},b\Delta m_t,c\Delta v_t)
\end{equation}
and follow the common future exactly.  Parameter-only interventions peak
early, first-moment interventions peak later and usually carry the largest
isolated response, and second-moment deviations strongly modulate the response
when coupled with $m$.  These hybrid counterfactuals can lie away from states
reached by ordinary AdamW; we use them to measure channel dependence and
interaction.

\begin{table*}[ht]
\centering
\caption{Complete post-shock state-channel intervention in the controlled stage of Experiment~1.
Entries are medians after first aggregating selected candidates within each of
the 32 independent systems.  Channel order is $(\theta,m,v)$.}
\label{tab:app_exp1_state_interventions}
\small
\setlength{\tabcolsep}{4pt}
\begin{tabular}{lc|rrr|rrr}
\toprule
& & \multicolumn{3}{c|}{Standard probe}
& \multicolumn{3}{c}{Anisotropic probe}\\
Channels & State & $M$ & ARE & $h^\star$ & $M$ & ARE & $h^\star$\\
\midrule
None & $000$ & 0 & 0 & 1.00 & 0 & 0 & 1.00\\
$\theta$ & $100$ & $1.80\!\times\!10^{-4}$ & $3.75\!\times\!10^{-3}$ & 6.50
& $1.43\!\times\!10^{-3}$ & $2.95\!\times\!10^{-2}$ & 6.00\\
$m$ & $010$ & $8.60\!\times\!10^{-4}$ & $1.91\!\times\!10^{-2}$ & 15.75
& $8.90\!\times\!10^{-3}$ & $1.71\!\times\!10^{-1}$ & 16.25\\
$v$ & $001$ & $2.73\!\times\!10^{-4}$ & $4.79\!\times\!10^{-3}$ & 25.25
& $2.47\!\times\!10^{-3}$ & $4.86\!\times\!10^{-2}$ & 27.25\\
$\theta,m$ & $110$ & $9.98\!\times\!10^{-4}$ & $2.19\!\times\!10^{-2}$ & 14.25
& $9.72\!\times\!10^{-3}$ & $1.94\!\times\!10^{-1}$ & 15.00\\
$\theta,v$ & $101$ & $2.74\!\times\!10^{-4}$ & $4.85\!\times\!10^{-3}$ & 18.75
& $2.64\!\times\!10^{-3}$ & $4.40\!\times\!10^{-2}$ & 22.50\\
$m,v$ & $011$ & $8.29\!\times\!10^{-4}$ & $1.68\!\times\!10^{-2}$ & 14.50
& $6.70\!\times\!10^{-3}$ & $1.30\!\times\!10^{-1}$ & 17.50\\
$\theta,m,v$ & $111$ & $9.23\!\times\!10^{-4}$ & $1.96\!\times\!10^{-2}$ & 14.00
& $7.59\!\times\!10^{-3}$ & $1.48\!\times\!10^{-1}$ & 14.50\\
\bottomrule
\end{tabular}
\end{table*}

In matched-first-displacement sweeps, the first future parameter displacement
is held fixed while $\beta_1$ or $\beta_2$ is varied.  Increasing $\beta_1$
from $0.5$ to $0.99$ moves the momentum-channel extremum from approximately
horizon 5 toward the end of the measured interval and substantially increases
ARE.  A horizon-$128$ extension confirms the ordering of time scales:
parameter deviations act earliest, momentum dominates intermediate delays,
and second-moment effects persist longest.  This rules out a purely larger
first-update explanation for the memory effect.

\begin{table*}[ht]
\centering
\caption{Controlled-stage matched-first-displacement persistence sweep in Experiment~1.  The
first future parameter displacement is held fixed within each channel.}
\label{tab:app_exp1_persistence}
\small
\setlength{\tabcolsep}{4pt}
\begin{tabular}{lc|rrr|rrr}
\toprule
& & \multicolumn{3}{c|}{Standard probe}
& \multicolumn{3}{c}{Anisotropic probe}\\
Channel & Decay & $M$ & ARE & $h^\star$ & $M$ & ARE & $h^\star$\\
\midrule
$m$ & 0.5  & $6.58\!\times\!10^{-5}$ & $1.08\!\times\!10^{-3}$ & 5.25
& $5.75\!\times\!10^{-4}$ & $1.08\!\times\!10^{-2}$ & 6.25\\
$m$ & 0.8  & $1.44\!\times\!10^{-4}$ & $2.74\!\times\!10^{-3}$ & 9.75
& $1.33\!\times\!10^{-3}$ & $2.61\!\times\!10^{-2}$ & 11.75\\
$m$ & 0.9  & $2.66\!\times\!10^{-4}$ & $5.96\!\times\!10^{-3}$ & 16.75
& $2.59\!\times\!10^{-3}$ & $5.46\!\times\!10^{-2}$ & 17.75\\
$m$ & 0.95 & $7.16\!\times\!10^{-4}$ & $1.47\!\times\!10^{-2}$ & 30.50
& $5.62\!\times\!10^{-3}$ & $1.20\!\times\!10^{-1}$ & 28.75\\
$m$ & 0.99 & $4.65\!\times\!10^{-3}$ & $7.23\!\times\!10^{-2}$ & 32.00
& $3.41\!\times\!10^{-2}$ & $5.34\!\times\!10^{-1}$ & 32.00\\
\midrule
$v$ & 0.9    & $7.82\!\times\!10^{-6}$ & $1.41\!\times\!10^{-4}$ & 9.00
& $7.31\!\times\!10^{-5}$ & $1.42\!\times\!10^{-3}$ & 10.50\\
$v$ & 0.99   & $1.37\!\times\!10^{-5}$ & $2.87\!\times\!10^{-4}$ & 17.75
& $1.33\!\times\!10^{-4}$ & $2.74\!\times\!10^{-3}$ & 17.50\\
$v$ & 0.999  & $4.64\!\times\!10^{-5}$ & $8.13\!\times\!10^{-4}$ & 24.75
& $4.31\!\times\!10^{-4}$ & $8.29\!\times\!10^{-3}$ & 26.75\\
$v$ & 0.9999 & $1.26\!\times\!10^{-4}$ & $2.24\!\times\!10^{-3}$ & 31.25
& $1.21\!\times\!10^{-3}$ & $1.83\!\times\!10^{-2}$ & 31.75\\
\bottomrule
\end{tabular}
\end{table*}

\subsubsection{Nonconvex Neural Networks}
\label{app:exp1_neural_details}
\label{app:exp2_details}

\paragraph{Models, data, and replication.}
We use CIFAR-10 with a two-hidden-layer width-256 MLP with GELU activations
($855{,}050$ parameters) and a three-layer width-32 CNN with ReLU activations
($94{,}538$ parameters).  Training batches contain 128 examples and fixed
test probes contain 256 examples.  Images use standard CIFAR-10 channel
normalization.  AdamW uses learning rate $2\times10^{-4}$,
$\beta_1=0.9$, $\beta_2=0.999$, $\epsilon=10^{-8}$, and weight decay
$0.01$; state and tangent calculations use double precision.

For each architecture, four top-level seeds (2026--2029) and four independent
initializations per seed give 16 independent systems, or 32 total.  Each
system receives 100 burn-in updates, four reference batches, 12 candidate
batches, 11 common-future batches, and one fixed probe, giving $H=12$ and
384 candidate shocks.  We use
$\alpha\in\{0.0625,0.125,0.25,0.5,1\}$.

\paragraph{Trajectory fidelity.}
Table~\ref{tab:app_exp2_fidelity} gives the complete finite-scale curve for the neural-network stage.  The
pooled median local NRMSE is $0.0678$ and median local sign agreement is one.
The median empirical validity radius is $0.75$ for CNN--ReLU (IQR
$[0.5,1]$) and $0.5$ for MLP--GELU (IQR $[0.375,0.5]$).  The recursive
state-error identity has maximum residual $9.33\times10^{-17}$.

\begin{table*}[ht]
\centering
\caption{Neural-network finite-horizon tangent fidelity in Experiment~1.  Entries are medians
over independent systems after aggregating candidates within systems.}
\label{tab:app_exp2_fidelity}
\small
\setlength{\tabcolsep}{5pt}
\begin{tabular}{lcccccc}
\toprule
Architecture & $\alpha$ & NRMSE & Sign agr. & Rel. $M$ err.
& Rel. ARE err. & Extremum-sign acc.\\
\midrule
CNN--ReLU & 0.0625 & 0.0458 & 1.000 & 0.0113 & 0.0115 & 1.000\\
& 0.125 & 0.0610 & 1.000 & 0.0160 & 0.0185 & 1.000\\
& 0.25  & 0.1097 & 1.000 & 0.0299 & 0.0307 & 1.000\\
& 0.5   & 0.1959 & 1.000 & 0.0587 & 0.0632 & 1.000\\
& 1    & 0.3854 & 0.958 & 0.1094 & 0.1126 & 1.000\\
\midrule
MLP--GELU & 0.0625 & 0.0377 & 1.000 & 0.0116 & 0.0140 & 1.000\\
& 0.125 & 0.0759 & 1.000 & 0.0230 & 0.0276 & 1.000\\
& 0.25  & 0.1530 & 1.000 & 0.0462 & 0.0566 & 1.000\\
& 0.5   & 0.2980 & 1.000 & 0.1002 & 0.1191 & 1.000\\
& 1    & 0.6294 & 0.917 & 0.1698 & 0.2270 & 1.000\\
\bottomrule
\end{tabular}
\end{table*}

\paragraph{Finite-scale error behavior.}
The local absolute-error exponent is fitted over
$\alpha\in\{0.0625,0.125,0.25\}$.  For MLP--GELU, its median remains in
$[2.001,2.005]$ at all horizons, with $R^2\approx1$.  For CNN--ReLU, the
median exponent is $2.017$, $1.916$, $1.573$, and $1.120$ at horizons
$1$, $4$, $8$, and $12$.  ReLU activation-sign differences are nonzero and
increase with scale (at $h=8$, their median rises from
$1.79\times10^{-4}$ at $\alpha=.0625$ to $2.82\times10^{-3}$ at
$\alpha=1$).  The degradation coincides with increasing activation-pattern changes,
consistent with a growing switching defect at later horizons.

\paragraph{Ranking future absolute influence.}
Within each system, we rank candidates by
$M_i=\max_h|d_{i,h}(1)|$.  Table~\ref{tab:app_exp2_ranking} reports the main
ablations for the neural-network stage.  Full ISO obtains pooled median Spearman $0.832$.  It exceeds the
exact one-step oracle and static norms in both architectures, whereas frozen
dynamics preserves much of the ordering over the short $H=12$ horizon.
Paired Wilcoxon tests use the independent system as the unit and Holm
correction within each architecture.  For CNN--ReLU, Full ISO significantly
exceeds all alternatives except frozen dynamics.  For MLP--GELU, its
advantages over exact one-step, gradient norm, parameter write, curvature,
norm product, and frozen readout survive correction; its differences from
the remaining propagation ablations are positive in median but not
significant after correction.

\begin{table}[ht]
\centering
\caption{Median within-system Spearman correlation with future magnitude in
the neural-network stage of Experiment~1.}
\label{tab:app_exp2_ranking}
\small
\setlength{\tabcolsep}{3pt}
\begin{tabular}{lcc}
\toprule
Method & CNN--ReLU & MLP--GELU\\
\midrule
Full ISO & \textbf{0.888} & \textbf{0.762}\\
Frozen dynamics & 0.843 & 0.671\\
Clamped parameter state & 0.818 & 0.661\\
No propagation & 0.755 & 0.668\\
Initial parameter only & 0.734 & 0.720\\
Frozen readout & 0.713 & 0.507\\
Exact one-step & 0.545 & 0.566\\
Gradient norm & 0.535 & 0.336\\
Parameter write norm & 0.490 & 0.283\\
Curvature & 0.559 & 0.248\\
Norm product & 0.241 & 0.224\\
\bottomrule
\end{tabular}
\end{table}

\paragraph{Readout order and signed extrema.}
Under 100 shared permutations of future readout order, CNN--ReLU Full ISO
decreases from $0.888$ to median $0.776$ ($p=0.0019$), with the true score
at the median 99th percentile of the system-specific null.  For MLP--GELU,
the corresponding values are $0.762$ and $0.713$ ($p=0.191$); its nearby
readouts are sufficiently similar that permutation preserves much of the
ordering.  Across the 96 exact intervention trajectories included in this diagnostic, 57 extrema are
positive and 39 are negative. A one-sided positive-response analysis would
therefore omit $40.6\%$ of these extremal events.

\paragraph{State-channel dependence.}
Table~\ref{tab:app_exp2_channels} reports exact hybrid-state interventions in the neural-network stage.
The parameter-only response peaks early, while the first-moment-only response
peaks later and has substantially greater accumulated magnitude.  The
second-moment-only response is small, but adding $v$ to $(\theta,m)$ slightly
reduces the median response in both architectures.  The conditional $v$
contrast is negative in 15 of 16 CNN systems and 13 of 16 MLP systems.  This
is evidence that $v$ modulates a coupled parameter--moment state, not that it
has a uniquely defined negative additive contribution.

\begin{table}[ht]
\centering
\caption{Neural-network exact state-channel interventions in Experiment~1.  Values are median
ARE relative to the full $111$ intervention; bit order is $(\theta,m,v)$.}
\label{tab:app_exp2_channels}
\small
\begin{tabular}{lcccc}
\toprule
State & CNN ARE & CNN $h^\star$ & MLP ARE & MLP $h^\star$\\
\midrule
$100$ & 0.216 & 2.5 & 0.250 & 1.5\\
$010$ & 0.821 & 9.0 & 0.827 & 7.5\\
$001$ & 0.030 & 10.0 & 0.015 & 9.5\\
$110$ & 1.025 & 9.0 & 1.010 & 6.5\\
$101$ & 0.214 & 3.0 & 0.237 & 1.5\\
$011$ & 0.812 & 9.0 & 0.818 & 7.5\\
$111$ & 1.000 & 9.0 & 1.000 & 7.0\\
\bottomrule
\end{tabular}
\end{table}

\paragraph{Matched-displacement persistence.}
Holding the first future parameter displacement fixed, increasing $\beta_1$
from $0.5$ to $0.99$ multiplies $M$ by $8.47$ (CNN) and $10.80$ (MLP),
and multiplies ARE by $14.17$ and $11.96$.  The median extremum moves from
approximately $h=3.5$--$4$ to $h=12$.  Increasing $\beta_2$ from $0.9$
to $0.9999$ multiplies ARE by $4.24$ and $2.75$, respectively.  Since the
first displacement is matched, these differences isolate persistence from
immediate update magnitude.

\subsubsection{Pretrained Language Models}
\label{app:exp1_llm_details}
\label{app:exp3_details}

\paragraph{Models, domains, and warm optimizer state.}
We use Pythia-410M, Pythia-1B, and Pythia-1.4B with WikiText-103,
OpenWebText, and CodeParrot, giving nine model--dataset systems.  From each
domain we prepare 2,048 token sequences of length 129.  The training batch
size is one and the fixed probe contains two sequences.  AdamW uses learning
rate $10^{-5}$, $\beta_1=0.9$, $\beta_2=0.999$, $\epsilon=10^{-8}$, and
weight decay $0.01$.  Starting from pretrained weights, every system is first
continued for 500 updates.  This creates a nontrivial trained AdamW moment
state and prevents a zero-moment cold start from dominating the derivative.
The intervention uses two reference batches, eight candidate shocks, seven
common-future batches, and $H=8$.  We evaluate
$\alpha\in\{0.0625,0.125,0.25,0.5,1\}$ under seed 2026.

\paragraph{Numerical ISO directional response.}
Directly differencing nearly equal scalar cross-entropies is inaccurate at
this scale.  For candidate $i$, horizon $h$, and probe example $n$, we
therefore compute the centered logit derivative
\begin{equation}
    \delta z_{i,h,n}(\varepsilon)
    =\frac{z^+_{i,h,n}(\varepsilon)-z^-_{i,h,n}(\varepsilon)}{2\varepsilon}
\end{equation}
and apply the exact control cross-entropy differential
\begin{equation}
    \widehat d_{i,h}(\varepsilon)
    =\frac{1}{N_{\mathrm{tok}}}
      \sum_n\left\langle
      \operatorname{softmax}(z^0_{h,n})-e_{y_n},
      \delta z_{i,h,n}(\varepsilon)\right\rangle.
\end{equation}
The resulting trajectory estimates the directional response
\begin{equation}
    \left.\frac{d}{d\alpha}
    F(\theta_{i,t+h}(\alpha))\right|_{\alpha=0},
\end{equation}
which equals $c_{t+h}^\top\Phi_{t+h,t+1}B_t\xi_i$ under the theorem's local
conditions.  We denote this numerical quantity by \textsc{ISO Tangent (FD)}.
Unlike the controlled and neural-network stages above, this stage estimates the end-to-end tangent
response without separately materializing $B_t$, every $A_s$, and the readout $c_{t+h}$.
The exact finite-scale target remains the FP64-reduced difference between the
shock and control probe losses.  We test
$\varepsilon\in\{.5,.25,.125,.0625,.03125,.015625\}$.  Adjacent derivative
trajectories are compared with the symmetric NRMSE
\begin{equation}
    E_{\mathrm{FD}}(\varepsilon,\varepsilon/2)
    =\frac{\|\widehat d(\varepsilon)-\widehat d(\varepsilon/2)\|_2}
    {\tfrac12(\|\widehat d(\varepsilon)\|_2+
    \|\widehat d(\varepsilon/2)\|_2)+10^{-30}}.
\end{equation}
The smaller scale in the most consistent adjacent pair is selected, and a
candidate is identifiable when the selected discrepancy is at most $0.25$.
All 72 candidates pass.  Median selected consistency is $0.00378$ and the
maximum accepted value is $0.22765$.  Selected smaller scales .25, .125,
.0625, .03125, and .015625 occur for 35, 15, 14, 4, and 4 candidates.

\paragraph{Finite-scale fidelity and validity.}
Table~\ref{tab:app_exp3_alpha} reports all shock scales.  The empirical
validity criterion is the median horizon-wise symmetric relative error at
most $0.2$.  Valid fractions are $93.1\%$, $97.2\%$, $87.5\%$, $73.6\%$,
and $55.6\%$ from the smallest to the largest scale.  The slight first-pair
nonmonotonicity reflects numerical variation around small responses.  The
result characterizes the finite-scale range over which the local mechanism
provides accurate pointwise predictions.

\begin{table}[ht]
\centering
\caption{Language-model finite-scale trajectory fidelity in Experiment~1 over 72 candidates.}
\label{tab:app_exp3_alpha}
\small
\setlength{\tabcolsep}{3pt}
\begin{tabular}{cccccc}
\toprule
$\alpha$ & NRMSE & 75\% NRMSE & Cosine & Rel. $M$ & Sign acc.\\
\midrule
0.0625 & 0.0387 & 0.1664 & 0.99990 & 0.0224 & 1.000\\
0.125  & 0.0477 & 0.1906 & 0.99993 & 0.0263 & 1.000\\
0.25   & 0.0946 & 0.2095 & 0.99983 & 0.0524 & 1.000\\
0.5    & 0.2105 & 0.4431 & 0.99938 & 0.1081 & 0.986\\
1     & 0.3902 & 0.7867 & 0.99668 & 0.2235 & 0.903\\
\bottomrule
\end{tabular}
\end{table}

\paragraph{Model- and domain-scale behavior.}
Table~\ref{tab:app_exp3_scale} shows no systematic local degradation from
410M to 1.4B.  Pooling scales, local NRMSE is $0.0426$ on CodeParrot,
$0.1094$ on OpenWebText, and $0.1161$ on WikiText-103.  At $\alpha=1$,
$87.5\%$, $37.5\%$, and $41.7\%$ of the candidates in these domains satisfy
the validity criterion, demonstrating that finite-scale range varies more by
domain than by model size in these conditions.  The overall median fitted
local error exponent is $1.921$.  Per-system exponents are
$(1.878,1.089,0.658)$ at 410M, $(1.988,1.916,1.936)$ at 1B, and
$(1.992,1.951,2.025)$ at 1.4B for CodeParrot, OpenWebText, and WikiText-103,
respectively.  Thus, the overall local trend is near quadratic, with the main
deviations concentrated in the 410M OpenWebText and WikiText-103 conditions.

\begin{table}[ht]
\centering
\caption{Language-model scaling summary in Experiment~1.  Ranking correlations are medians over
the three data-domain conditions at each scale.}
\label{tab:app_exp3_scale}
\small
\setlength{\tabcolsep}{3pt}
\begin{tabular}{cccccc}
\toprule
Scale & Local NRMSE & Cosine & Valid 0.25 & Valid 1 & ISO-FD $\rho$\\
\midrule
0.41B & 0.1090 & 0.99918 & 0.958 & 0.417 & 0.714\\
1.0B  & 0.0519 & 0.99995 & 0.792 & 0.667 & 0.833\\
1.4B  & 0.0567 & 0.99994 & 0.875 & 0.583 & 0.762\\
\bottomrule
\end{tabular}
\end{table}

\paragraph{Ranking at full shock scale.}
The exact one-step baseline is
\begin{equation}
    S_i^{\mathrm{1step}}=|d_{i,1}(1)|,
\end{equation}
which executes the complete nonlinear shock and control updates and evaluates
both on the fixed probe, but sees no subsequent common-future propagation.
It is therefore a strong counterfactual oracle that directly observes the
immediate fixed-probe effect.  Table~\ref{tab:app_exp3_ranking} reports the
system-wise results.
\textsc{ISO Tangent (FD)} is positive in all nine conditions and exceeds
gradient and parameter-write norms in eight.  It exceeds exact one-step in
five conditions, while
exact one-step is higher in four; their medians are equal.  The two statistics
capture complementary information: one-step observes the immediate output,
whereas \textsc{ISO Tangent (FD)} represents the propagated signed trajectory.

\begin{table*}[ht]
\centering
\caption{Language-model within-system Spearman correlation in Experiment~1 with
$M_i=\max_h|d_{i,h}(1)|$.  Each condition contains eight candidates.}
\label{tab:app_exp3_ranking}
\small
\begin{tabular}{llrrrr}
\toprule
Model & Dataset & ISO tangent (FD) & Exact 1-step & Gradient norm & Parameter write\\
\midrule
410M & CodeParrot   & 0.857 & 0.643 & -0.286 & -0.262\\
410M & OpenWebText  & 0.548 & 0.738 &  0.571 &  0.595\\
410M & WikiText-103 & 0.714 & 0.643 &  0.571 &  0.619\\
1B   & CodeParrot   & 0.952 & 0.905 & -0.167 & -0.048\\
1B   & OpenWebText  & 0.333 & 0.929 &  0.071 & -0.286\\
1B   & WikiText-103 & 0.833 & 0.762 & -0.214 &  0.048\\
1.4B & CodeParrot   & 0.976 & 0.905 & -0.167 & -0.167\\
1.4B & OpenWebText  & 0.762 & 0.929 &  0.429 &  0.405\\
1.4B & WikiText-103 & 0.619 & 0.667 & -0.119 &  0.190\\
\midrule
Median & --- & 0.762 & 0.762 & -0.119 & 0.048\\
\bottomrule
\end{tabular}
\end{table*}

As a descriptive secondary analysis, ranks normalized within each condition
give pooled correlations $0.791$ for one-step and $0.733$ for ISO-FD.  After
linearly controlling one-step rank, the partial ISO-FD--target correlation is
$0.310$; adding ISO-FD rank raises descriptive $R^2$ from $0.626$ to $0.662$.
Because candidates are nested within nine fixed conditions, these values are
descriptive and are not treated as 72 independent replications.

The exact one-step response is a strong baseline in the pretrained-language-model
setting and exceeds ISO-FD in several conditions. This does not contradict the
finite-horizon mechanism: the one-step-matched controlled experiments show
that immediate response is not sufficient in general, while the present
scaling experiment tests whether the ISO tangent remains faithful at larger
model scale.

\paragraph{Signed extrema.}
At full scale, 51 of 72 candidate extrema are positive and 21 are negative.
ISO-FD recovers the extremum sign for 65 candidates ($90.3\%$).  Counts by domain
are $(24,0)$ for CodeParrot, $(13,11)$ for OpenWebText, and $(14,10)$ for
WikiText-103, where each pair is (positive, negative).  This domain dependence
reinforces the signed formulation: positive loss excursions are one
subclass of the broader finite-horizon response.

\paragraph{State-channel interactions.}
For the single candidate selected per system in this diagnostic, we evaluate the same
eight post-shock hybrid states used in the smaller experiments.  The most
striking behavior occurs on WikiText-103: injecting $m$ without its matched
$v$ produces maximum absolute responses $50.322$, $8.5748$, and $26.339$ at
410M, 1B, and 1.4B, whereas injecting the coupled $(m,v)$ deviations gives
$0.01084$, $0.00886$, and $0.00847$.  Full-state responses are similarly of
order $10^{-2}$.  These extreme isolated-momentum hybrids demonstrate strong
cross-channel coupling and off-trajectory sensitivity.  We therefore
interpret the intervention family jointly as an interaction probe.

\paragraph{Computational cost and scope.}
Median complete runtime per system, including 500 continuation updates,
eight shocks, the finite-difference grid, exact trajectories, and channel
interventions, is approximately 464 seconds at 410M, 645 seconds at 1B, and
941 seconds at 1.4B.  The nine measured runtimes sum to approximately 5,849
seconds and were parallelized over four GPUs.  With one seed, this experiment
provides a descriptive scaling comparison over three sizes of the Pythia
family and three data domains.  Multi-seed replication of the same mechanism
is provided by the controlled and neural-network stages of Experiment~1.

\subsection{Experiment 2: Prospective Structure Under Unknown Futures}
\label{app:exp2_predictability}

\paragraph{Repeated-future construction.}
Experiment~1 conditions on one realized common future.  Experiment~2 instead
holds the pre-shock history, post-burn-in AdamW state, reference minibatches,
candidate shock, fixed probe, optimizer configuration, and horizon fixed, and
resamples only the unseen future minibatches.  For each candidate $i$ and
future branch $k\in\{1,\ldots,K\}$, the control and shock trajectories share
the same branch, so that
\begin{equation}
    d_{i,h}^{(k)}
    =
    F(\theta_{i,t+h}^{\mathrm s,(k)})
    -
    F(\theta_{t+h}^{\mathrm c,(k)})
\end{equation}
continues to isolate the response to the initiating shock within that branch.
We use $K=32$ throughout.

In the controlled systems, the original burn-in state, four reference
minibatches, 16 candidates, probes, and $H=32$ are unchanged.  The latent
state of the correlated quadratic process immediately before the future is
held fixed and each branch resamples only subsequent innovations.  One branch
is shared across all candidates and both probe views in a base system.  The
\texttt{standard} and \texttt{readout\_active} results below correspond to the
standard and anisotropic probes from Experiment~1; the rotating-readout probe
is not used in this repeated-future experiment.

In the neural systems, the original CIFAR-10 architecture, initialization,
100-step burn-in state, four reference minibatches, 12 candidates, fixed test
probe, AdamW configuration, and $H=12$ are unchanged.  Each future branch
independently resamples the subsequent CIFAR-10 training minibatches and is
shared across candidates and candidate modes within a system.

\paragraph{Candidate families and response summaries.}
Both settings evaluate the original \emph{natural} gradient residuals and an
\emph{exact-one-step-matched} family.  The matched family reuses the
construction in Eq.~\eqref{eq:app_exact_matching}: the common tangent target is
$0.25$ times the median natural one-step magnitude, the candidate-specific
component is placed in the nullspace of the immediate output functional with
four times the norm of the common component, and one nonnegative scalar per
candidate is calibrated through the exact nonlinear AdamW update.  No
response at $h>1$ and no resampled future branch is used in this construction.
All candidates calibrate successfully.  Table~\ref{tab:app_exp2_matching_audit}
shows that the resulting exact $|d_1|$ values are constant to numerical
precision.

For each candidate--branch pair we record
\begin{align}
    M_{i,k}
    &:=\max_h|d_{i,h}^{(k)}|,\\
    P^+_{i,k}
    &:=\max_h[d_{i,h}^{(k)}]_+,\\
    A_{i,k}
    &:=\sum_h|d_{i,h}^{(k)}|.
\end{align}
Magnitude $M$ is the primary response used in the main text; $P^+$ and $A$
are secondary checks.  For a generic summary $R_{i,k}$, define
\begin{equation}
    \mu_i^R:=\frac1K\sum_{k=1}^K R_{i,k},
\end{equation}
and the within-system prospective-structure ratio
\begin{equation}
\label{eq:app_predictability_ratio}
    \Pi_H(R)
    =
    \frac{
      \operatorname{Var}_i(\mu_i^R)
    }{
      \operatorname{Var}_i(\mu_i^R)
      +
      \mathbb E_i[\operatorname{Var}_k(R_{i,k})]
    }.
\end{equation}
The implementation uses population normalization within the finite candidate
and branch grids.  $\Pi_H$ is therefore a protocol-specific variance
decomposition: it compares candidate-specific variation in the repeated-future
mean with variation induced by resampling the future, and is not an
information-theoretic percentage or a universal fraction of predictable risk.
We additionally compute
\begin{equation}
    \rho_{\mathrm{branch}}
    :=
    \operatorname{median}_k
    \operatorname{Spearman}(R_{\cdot,k},\mu^R),
\end{equation}
which measures the stability of candidate ordering across individual future
branches.

\paragraph{Present-time scores.}
The ranking target for a present-time score is the repeated-future conditional
mean $\mu_i^M$.  The scalar baselines are the exact one-step absolute response,
gradient norm, and initial parameter-write norm from
Eq.~\eqref{eq:app_scalar_baselines}.  We add two ISO-based scores that are
computed without accessing any sampled future branch.  The
\emph{present-frozen ISO} evaluates one local transition Jacobian
$\overline A_t$ at the first post-control state using only the current
reference minibatches, freezes the $h=1$ probe readout $\bar c_1$, and uses
\begin{equation}
\label{eq:app_present_frozen_iso}
    S_i^{\mathrm{PF}}
    =
    \max_{1\leq h\leq H}
    \left|
      \bar c_1^\top
      P_\theta
      \overline A_t^{\,h-1}
      B_t\xi_i
    \right|.
\end{equation}
The \emph{reference-surrogate ISO} instead constructs a deterministic future
using only the current reference-minibatch mean (controlled systems) or the
current reference minibatches (neural systems), updates the resulting
surrogate control state, readout, and tangent along that deterministic
rollout, and takes the maximum absolute predicted response.  The distinction
from the pathwise Full ISO in Experiment~1 is essential: neither prospective
score uses a realized or resampled future minibatch.

\subsubsection{Controlled Repeated Futures}
\label{app:exp2_controlled_predictability}

Table~\ref{tab:app_exp2_predictability_primary} reports the primary magnitude
results.  Under the standard probe, natural candidates have median
$\Pi_{32}(M)=0.801$, but the exact one-step response already correlates
$0.801$ with $\mu_i^M$.  After exact one-step matching,
$\Pi_{32}(M)=0.730$ and the median branch-ranking correlation remains $0.909$,
despite the one-step magnitude being tied to numerical precision.  The
anisotropic probe gives the same qualitative conclusion, with matched
$\Pi_{32}(M)=0.603$ and branch correlation $0.895$.  In contrast, the simple
present-time scores do not recover the matched conditional-mean ordering in
this controlled setting: present-frozen ISO correlations are $-0.066$ and
$0.031$ for the standard and anisotropic probes.

\subsubsection{Neural-Network Repeated Futures}
\label{app:exp2_neural_predictability}

The neural results preserve the same protocol while replacing the controlled
quadratic process by independently resampled CIFAR-10 future minibatches.
Matched MLP--GELU yields $\Pi_{12}(M)=0.921$ with branch-ranking correlation
$0.955$; matched CNN--ReLU yields $0.700$ and $0.872$.  Present-time ISO
scores are substantially more informative here than in the controlled
systems.  Present-frozen ISO correlates $0.941$ with $\mu_i^M$ for MLP--GELU
and $0.755$ for CNN--ReLU, while reference-surrogate ISO reaches $0.752$ and
$0.811$.  These results establish prospective relevance under the measured
protocols, but they do not by themselves identify the conditions that make a
present-time representation accurate.

\begin{table*}[ht]
\centering
\caption{Primary repeated-future magnitude results in Experiment~2.
Each cell reports the median across independent systems; brackets give the
system-level IQR.  Branch $\rho$ is the within-system median correlation
between an individual future-branch ranking and the repeated-future mean
ranking.}
\label{tab:app_exp2_predictability_primary}
\small
\setlength{\tabcolsep}{3.8pt}
\begin{tabular}{llcc}
\toprule
System & Candidates & $\Pi_H(M)$ & Branch $\rho$\\
\midrule
Quadratic, standard
 & Natural & $0.801\ [0.709,0.877]$ & $0.919\ [0.866,0.949]$\\
 & Matched & $0.730\ [0.594,0.845]$ & $0.909\ [0.839,0.941]$\\
Quadratic, anisotropic
 & Natural & $0.718\ [0.602,0.825]$ & $0.899\ [0.846,0.943]$\\
 & Matched & $0.603\ [0.501,0.727]$ & $0.895\ [0.844,0.931]$\\
MLP--GELU
 & Natural & $0.771\ [0.621,0.844]$ & $0.862\ [0.795,0.914]$\\
 & Matched & $0.921\ [0.885,0.951]$ & $0.955\ [0.937,0.959]$\\
CNN--ReLU
 & Natural & $0.716\ [0.649,0.758]$ & $0.855\ [0.823,0.900]$\\
 & Matched & $0.700\ [0.595,0.771]$ & $0.872\ [0.855,0.892]$\\
\bottomrule
\end{tabular}
\end{table*}

\begin{table*}[ht]
\centering
\caption{Present-time Spearman correlation with the repeated-future mean
magnitude $\mu_i^M$ in Experiment~2.  The exact one-step score is tied in the
matched family and its correlation is therefore undefined.}
\label{tab:app_exp2_present_scores}
\small
\setlength{\tabcolsep}{4.2pt}
\begin{tabular}{llrrrrr}
\toprule
System & Candidates & Exact 1-step & Grad. norm & Write norm
& Frozen ISO & Ref.-surrogate ISO\\
\midrule
Quadratic, standard
 & Natural & 0.801 & 0.212 & 0.253 & 0.659 & 0.372\\
 & Matched & -- & 0.049 & -0.057 & -0.066 & -0.021\\
Quadratic, anisotropic
 & Natural & 0.790 & 0.394 & 0.374 & 0.562 & 0.490\\
 & Matched & -- & 0.272 & 0.025 & 0.031 & 0.025\\
MLP--GELU
 & Natural & 0.661 & 0.318 & 0.325 & 0.734 & 0.262\\
 & Matched & -- & 0.066 & 0.339 & 0.941 & 0.752\\
CNN--ReLU
 & Natural & 0.619 & 0.538 & 0.510 & 0.752 & 0.601\\
 & Matched & -- & 0.500 & 0.500 & 0.755 & 0.811\\
\bottomrule
\end{tabular}
\end{table*}

\begin{table*}[ht]
\centering
\caption{Exact-one-step matching audit for Experiment~2.  CV is computed
within each candidate family; entries are medians with IQRs.  Calibration
success is 1.000 in every condition.}
\label{tab:app_exp2_matching_audit}
\small
\setlength{\tabcolsep}{4pt}
\begin{tabular}{lc}
\toprule
Condition & CV of $|d_1|$\\
\midrule
Quadratic, standard
& $1.07\!\times\!10^{-12}\ [7.97\!\times\!10^{-13},1.75\!\times\!10^{-12}]$\\
Quadratic, anisotropic
& $1.33\!\times\!10^{-12}\ [6.50\!\times\!10^{-13},1.84\!\times\!10^{-12}]$\\
MLP--GELU
& $3.57\!\times\!10^{-13}\ [2.79\!\times\!10^{-13},5.73\!\times\!10^{-13}]$\\
CNN--ReLU
& $1.36\!\times\!10^{-12}\ [1.08\!\times\!10^{-12},2.02\!\times\!10^{-12}]$\\
\bottomrule
\end{tabular}
\end{table*}

\begin{table*}[ht]
\centering
\caption{Absolute variance components for the matched primary magnitude.
Values are medians of the system-specific quantities in
Eq.~\eqref{eq:app_predictability_ratio}; the ratio of the two medians need not
equal the median of the system-specific ratios.}
\label{tab:app_exp2_variance_components}
\small
\setlength{\tabcolsep}{5pt}
\begin{tabular}{lcc}
\toprule
Condition & Between-candidate variance & Mean within-future variance\\
\midrule
Quadratic, standard    & $9.92\!\times\!10^{-10}$ & $2.39\!\times\!10^{-10}$\\
Quadratic, anisotropic & $4.25\!\times\!10^{-8}$  & $2.35\!\times\!10^{-8}$\\
MLP--GELU              & $4.31\!\times\!10^{-8}$  & $5.45\!\times\!10^{-9}$\\
CNN--ReLU              & $2.02\!\times\!10^{-8}$  & $1.13\!\times\!10^{-8}$\\
\bottomrule
\end{tabular}
\end{table*}

\begin{table*}[ht]
\centering
\caption{Prospective-structure ratio for the primary magnitude and two
secondary response summaries.  $P^+$ is sign-sensitive; because the matched
construction enforces alternating signed one-step targets, the positive-peak
column is reported only as a descriptive robustness check and is not used to
support the main prospective-identifiability claim.}
\label{tab:app_exp2_secondary_metrics}
\small
\setlength{\tabcolsep}{5pt}
\begin{tabular}{llccc}
\toprule
System & Candidates & $\Pi_H(M)$ & $\Pi_H(A)$ & $\Pi_H(P^+)$\\
\midrule
Quadratic, standard
 & Natural & 0.801 & 0.817 & 0.704\\
 & Matched & 0.730 & 0.729 & 0.995\\
Quadratic, anisotropic
 & Natural & 0.718 & 0.722 & 0.570\\
 & Matched & 0.603 & 0.603 & 0.990\\
MLP--GELU
 & Natural & 0.771 & 0.816 & 0.807\\
 & Matched & 0.921 & 0.887 & 0.995\\
CNN--ReLU
 & Natural & 0.716 & 0.738 & 0.787\\
 & Matched & 0.700 & 0.791 & 0.940\\
\bottomrule
\end{tabular}
\end{table*}

The combined result is deliberately narrower than an online-prediction claim.
Repeated futures show that candidate identity can remain a strong source of
finite-horizon variation even after the immediate output magnitude is matched,
while the success of a particular present-time ISO representation varies
substantially across systems.  Experiment~2 therefore motivates prospective
identifiability as a separate question without assuming that the pathwise ISO
itself is deployable before the future trajectory is observed.

\end{document}